\documentclass[aoas]{imsart}
\usepackage{bm}
\usepackage{xcolor}
\usepackage{bbold}
\usepackage{multirow}
\usepackage{booktabs} 
\usepackage{float}
\usepackage[para,online,flushleft]{threeparttable}

\usepackage{lipsum}

\RequirePackage{amsthm,amsmath,amsfonts,amssymb}
\RequirePackage[authoryear]{natbib}

\RequirePackage[colorlinks,citecolor=blue,urlcolor=blue]{hyperref}
\RequirePackage{graphicx}

\startlocaldefs
\theoremstyle{plain}

\newtheorem{theorem}{Theorem}[section]

\theoremstyle{definition}

\newtheorem{remark}{Remark}

\newtheorem{proposition}{Proposition}

\usepackage{algorithm}
\usepackage{algpseudocode}

\newcommand{\Qmat}{\bm{W}_{s, l}^{(Q)}}
\newcommand{\Kmat}{\bm{W}_{s, l}^{(K)}}
\newcommand{\Vmat}{\bm{W}_{s, l}^{(V)}}
\newcommand{\Promat}{\bm{W}_{s, l}^{(O)}}
\newcommand{\MLPsec}{\bm{W}_{l}^{(2)}}
\newcommand{\MLPfir}{\bm{W}_{l}^{(1)}}
\newcommand{\MLPBiassec}{\bm{b}_{l}^{(2)}}
\newcommand{\MLPBiasfir}{\bm{b}_{l}^{(1)}}

\endlocaldefs

\begin{document}

\begin{frontmatter}
\title{Copula-enhanced Vision Transformer for high myopia diagnosis through OU UWF fundus images}
\runtitle{Copula enhanced ViT}

\begin{aug}
\author[A]{\fnms{Chong}~\snm{Zhong}},
\author[B]{\fnms{Liu}~\snm{Yunhao}}, 
\author[C]{\fnms{Yang}~\snm{Li}}, 
\author[C]{\fnms{Fu}~\snm{Xiang}}, 
\author[D]{\fnms{Jin}~\snm{Yang}}
\author[F]{\fnms{Danjuan}~\snm{Yang}},
\author[F]{\fnms{Meiyan}~\snm{Li}},
\author[E]{\fnms{Xu}~\snm{Jinfeng}}, 
\author[G]{\fnms{Liu}~\snm{Aiyi}}, 
\author[H]{\fnms{A.H.}~\snm{Welsh}}, 
\author[F]{\fnms{Xingtao}~\snm{Zhou}},
\author[C]{\fnms{Bo}~\snm{Fu$^{\ddag}$}\ead[label=e1]{fu@fudan.edu.cn}}, 
\and
\author[B]{\fnms{Catherine C.}~\snm{Liu$^{\ddag}$}\ead[label=e2]{macliu@polyu.edu.hk}
}

\address[A]{Department of Applied Biology and Chemical Technology, The Hong Kong Polytechnic University}

\address[B]{Department of Data Science \& AI, The Hong Kong Polytechnic University \printead[presep={,\ }]{e2}}

\address[C]{School of Data Science, Fudan University \printead[presep={,\ }]{e1}}

\address[D]{Department of Applied Mathematics, The Hong Kong Polytechnic University}

\address[E]{Department of Biostatistics, City University of Hong Kong}

\address[F]{Eye Institute and Department of Ophthalmology, Eye \& ENT Hospital, Fudan University}

\address[G]{Eunice Kennedy Shriver National Institute of Child Health and Human Development}

\address[H]{College of Business and Economics, Australian National University}

 \end{aug}

\begin{abstract}
The advancement of AI-assisted myopia screening necessitates the joint diagnosis of both-eye (OU) high myopia (HM) status and the prediction of axial length (AL).
This clinical requirement introduces a complex mixed-type (binary-continuous) multitask learning task with bi-domain (OU) image covariates, giving rise to two key challenges: i) capture the inter-ocular asymmetry of OU images within a cutting-edge foundation model; ii) model and estimate the conditional dependence structure among mixed-type multivariate responses given image covariates.
We address the challenges by: i) imposing residual adapters on the Vision Transformer foundation model to capture the OU similarity and heterogeneity simultaneously; 
ii) developing a four-dimensional copula loss that is implementable in PyTorch based on a latent variable expression for the Gaussian copula likelihood, and proposing a computationally efficient fast Monte Carlo Expectation Maximization (fMCEM) algorithm to estimate copula parameters. 
We further formulate a specific overfitting problem called stronger covariance phenomenon in multitask learning. 
We reveal the disturbance of the phenomenon to estimation of copula parameters and theoretically demonstrate the numerical stability of the proposed fMCEM algorithm against the disturbance. 
The application to our annotated OU ultra-widefield fundus image dataset and simulation on synthetic data demonstrate that our method stably enhances the predictive capabilities on both classification and regression tasks. \\

\end{abstract}

\begin{keyword}
\kwd{Copula model}
\kwd{EM algorithm}
\kwd{High myopia}
\kwd{Overfitting}
\kwd{Ultra-widefield fundus image}
\kwd{Vision Transformer}
\end{keyword}
\end{frontmatter}



\section{Introduction}
\label{sec:intro}
High myopia (HM) is a severe form of myopia that is commonly defined as a spherical equivalent refractive error of -6.0 diopters or less \citep{kobayashi2005fundus}. 
HM is associated with axial elongation and thinning of posterior ocular tissues, including the choroid, sclera, and retina \citep{jonas2023imi}. 
These changes increase the risk of vision-threatening complications such as retinal detachment, glaucoma, and myopic maculopathy, which may lead to irreversible visual impairment \citep{iwase2006prevalence}.
Consequently, early diagnosis and regular monitoring of HM are of great importance in ophthalmic care.

In ophthalmology research, there is a global trend to leverage artificial intelligence (AI) for the automated diagnosis of ophthalmological cases or diseases based on ocular images \citep{grzybowski2024retina}.
A set of AI models for myopia screening based on ocular images have emerged since 2020 \citep[among others]{choi2021deep, yang2022prediction, li2022advances, yoo2022deep, yew2025deep}, including two of our previous investigations, CeCNN \citep{zhong2025cecnn} and OU-Copula \citep{li2024oucopula}.  
Despite this rapid progress, existing approaches still face several challenges in meeting the complex needs of real-world clinical practice. To address these limitations, this paper integrates novel statistical modeling with cutting-edge AI techniques to provide a more comprehensive framework for AI-assisted myopia screening.

\noindent{\textbf{Motivation 1} (medical AI architecture):} new AI architecture to capture the similarity and heterogeneity of both-eye (OU, oculus uterque) ocular images simultaneously.  
Most of the existing models are ``single-eye" models in that they feed the ocular images of left and right eyes into a same convolutional neural network (CNN) without discrimination \citep{engelmann2022detecting}. 
However, clinical literature has pointed out the existence of ``inter-ocular asymmetry" i.e. the asymmetrical or
unilateral features between OU \citep{lu2022interocular}. 
On the other hand, OU ocular images generally have high similarity from the distributional perspective. 
We illustrate this point via the t-distributed Stochastic Neighbor Embedding \citep[t-SNE,][]{van2008visualizing}, a commonly used non-linear machine learning algorithm to visualize high-dimensional data (such as ocular images) on a 2D plane.

\begin{figure}[tb]
    \centering
    \includegraphics[scale = .2]{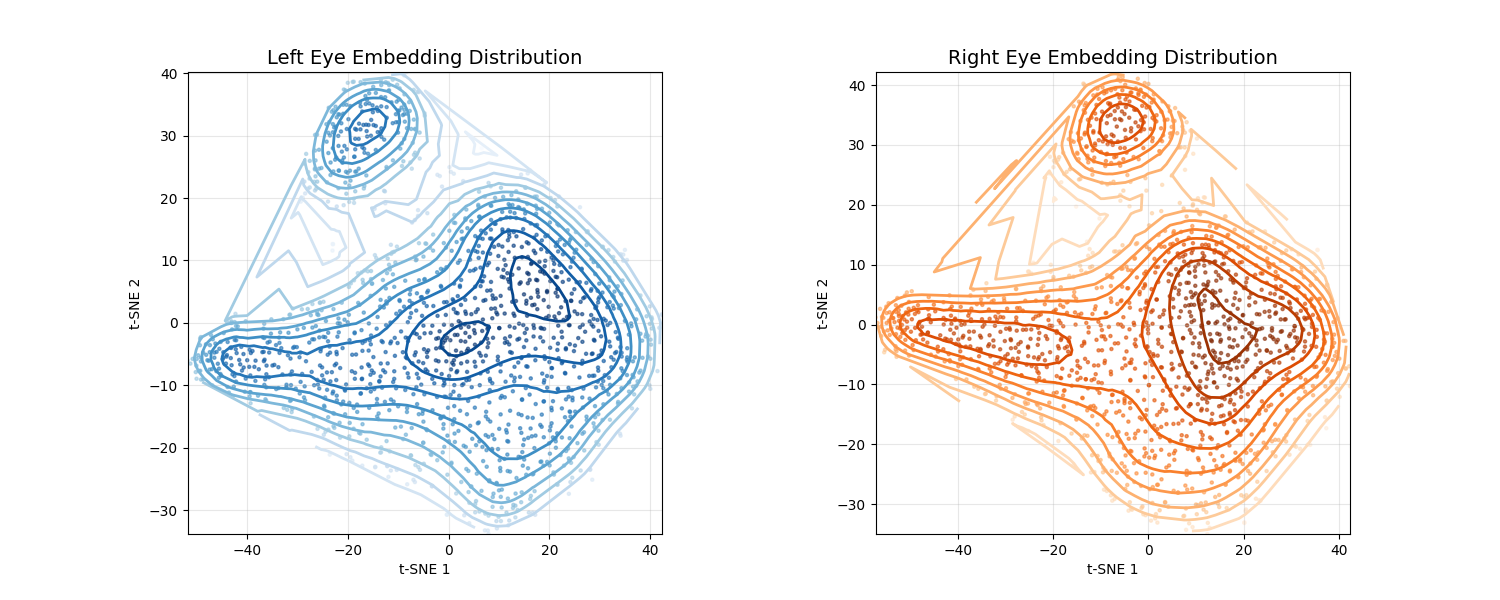}
    \caption{Contour plot of t-SNE of OU UWF fundus images; embeddings are extracted by Vision Transformer. }
    \label{fig:tsne_oueye}
\end{figure}

Figure \ref{fig:tsne_oueye} displays the contour plots of t-SNE of OU ultra-widefield (UWF) fundus image samples respectively, where the OU t-SNEs share almost identical distributions, except for slight differences on the shape of valleys. 
Given the high similarity between OU, it is inappropriate to feed the OU UWF fundus images into two separate neural networks, as it will lose half of the training sample size. 
As a pilot study, OU-Copula develops a bi-channel CNN to incorporate the inter-ocular asymmetry into a unified CNN for OU images and enhances the predictions of two OU continuous clinical scores, Axial length (AL) and spherical Equivalent (SE). 
However, in most computer vision tasks, Vision Transformer \citep[ViT,][]{dosovitskiy2020image} and its variants are state-of-the-art (SOTA) backbones and generally outperform CNNs. 
To this end, we develop a new pair-eye architecture for ViT and employ it on our UWF fundus image dataset.

\noindent{\textbf{Motivation 2} (applied statistics in AI): } new statistical modeling for joint predictions of OU HM (binary) and AL (continuous).
In the single-eye scenario, CeCNN has demonstrated that modeling the bivariate conditional dependence between AL and HM can enhance their predictions. 
In the OU setting, this problem becomes multivariate mixed-type  response nonlinear regression, yielding two issues in the field of statistics for AI. 
\begin{enumerate}
    \item[(i)] \textit{Computational statistics issue}: how to implement a loss function that incorporates the mixed-type conditional dependence structure in the Python package PyTorch \citep{paszke2019PyTorch}, the commonly used deep learning library. 
    \item[(ii)] \textit{Statistical inference issue}: how to estimate the conditional dependence structure given OU UWF fundus images based on neural network outputs. 
\end{enumerate}
 
For issue (i), the Gaussian copula model \citep{joe1997multivariate, Song2000multivariate} is widely used to model multivariate continuous-discrete data, with the general form of the mixed-type likelihood given by \cite{song2009joint}. 
Nonetheless, this general likelihood form needs to compute complicated multivariate Gaussian integrals, which are NOT implementable in PyTorch. 
If one employs other libraries such as SciPy to compute the likelihood, then the non differentiable nature of SciPy will hinder backpropagation in neural network training. 
To this end, we need an equivalent expression of the mixed-type likelihood that is feasible in PyTorch.

Issue (ii) under a Gaussian copula model is equivalent to estimating the correlation matrix in the Gaussian copula, i.e. $\Gamma$ in \eqref{Gamma matrix} of Section 2.2. 
For mixed-type data, the widely used approach is to draw the posterior of $\Gamma$ via Markov Chain Monte Carlo (MCMC) based on the extended rank-likelihood \citep{hoff2007extending, murray2013bayesian, cui2019novel}, which has been extended to regression scenarios with online learning algorithm \citep{alexopoulos2021bayesian, feldman2022bayesian}. 
However, in the deep learning case, online algorithms are intractable; refer to the discussion part in \cite{zhong2025cecnn}. 
If we adopt an offline algorithm deploying plug-in warm-up neural network estimators, the MCMC approaches may be problematic: we empirically observe that 
the MCMC estimates lead to either computational overflow or poor predictive capabilities, not to mention that MCMC approaches suffer from heavy computational burden on large-scale data; refer to Supplement B.2.  
The numerical instability of the MCMC approach may be a consequence of the overfitting problem; refer to Proposition \ref{prop: overestimation}. 
Although the Expectation Maximization (EM) algorithm is an efficient alternative to MCMC, the existing EM methods are generally applied to purely continuous responses \citep{ding2016algorithm, li2019expectation}. 
For Gaussian copula on mixed-type data, the EM algorithm has rarely been applied because the updating formulas in the E-step and M-step are both intractable. 
To this end, we develop a computationally efficient method to estimate the complicated mixed-type dependence structure accounting for the general overfitting problem of deep learning methods. 

In this paper, we propose the Copula-enhanced Vision Transformer (CeViT), a novel ophthalmology AI that jointly predicts OU HM and AL based on OU UWF fundus images. 
We impose dynamically weighted OU residual adapters (refer to Eq. \eqref{adapters}) on the original ViT encoder shared by both eyes. 
The shared encoder automatically learns the similarity between OU UWF fundus images as dominating features, and the two adapters learn the OU heterogeneity respectively as a complement. 
We present an implementable copula loss in PyTorch via a latent variable expression for the Gaussian copula likelihood (Eq. \eqref{joint log density}), avoiding the computation of complicated multivariate Gaussian integrals. 
We propose a fast Monte Carlo Expectation Maximization (fMCEM) algorithm to estimate the copula correlation matrix. 
The proposed algorithm transfers sampling procedures to expectation computation and enjoys fast convergence, enabling efficient computation on large-scale data. 
We formulate a two-stage training framework (Algorithm \ref{alg: CeViT}) that contains two training modules and an estimation module to facilitate stable and efficient training under the proposed copula loss.  

The overfitting problem is inevitable in deep learning practice since neural networks are often trained to perfectly fitting the training data \citep{zhang2016understanding}, let alone training the overparameterized ViT model on our UWF fundus image training set with limited data size. 
We report a specific overfitting problem called the \textit{stronger covariance phenomenon}, in the sense that
the Gaussian scores of a pair outputs of neural networks have larger covariance than the ground truth. 
This phenomenon is coincided with the random feature modeling of benign overfitting \citep{li2023benign}. 
Unlike existing discussions about overfitting that focus on generalization error behaviors \citep{bejani2021systematic}, 
we study how this overfitting phenomenon influences the estimation of the copula parameters in our multitask learning situation. 
Specifically, in the presence of the stronger covariance phenomenon, we reveal the overestimation and underestimation of copula parameter estimation on tasks of classification (Proposition \ref{prop: overestimation}) and regression (Proposition \ref{prop: underestimation regression}), respectively. 
On the one hand, the overestimation on classification tasks yields nearly singular MCMC estimates, explaining the numerical instability problem. 
On the other hand, we demonstrate that the proposed fMCEM algorithm balances the overestimation and consequently, ensures stable computation against overfitting (Proposition \ref{prop: no over 1}).

The remainder of the paper is organized as follows. 
Section \ref{sec: cevit architecture} presents the architecture of CeViT, the copula loss, and the fMCEM algorithm. 
Section \ref{sec: feasible} introduces the training framework and prediction scheme. 
Section \ref{sec: overfitting} characterizes the stronger covariance phenomenon and the disturbance. 
Sections \ref{sec:simulation} and \ref{sec: app} apply the CeVit to synthetic data and the annotated OU UWF fundus image data collected by The Eye \& ENT Hospital of Fudan University, respectively. 
Section \ref{sec: discussion} concludes the paper with brief discussions. 
Technical proofs and additional simulations are collected in the Supplementary materials. 
The code for simulations is available on GitHub \hyperlink{https://github.com/silent618/CeViT}{https://github.com/silent618/CeViT}.

\section{CeViT: architecture and loss}
\label{sec: cevit architecture}
Let $\bm{y} = (y_{1}, y_{2}, y_{3}, y_{4})^{{T}}$ be the 4-dimensional vector of mixed discrete-continuous responses, where $y_{1}, y_{2} \in \mathbb{R}$ correspond to AL of left and right eyes respectively, and $y_{3}, y_{4} \in \{0, 1\}$ correspond to the myopia status (1: high-myopia; 0: else) of left and right eyes respectively. 
Let $\mathcal{X}_j \in \mathbb{R}^{224 \times 224 \times 3}$ be the UWF fundus images of left and right eyes respectively, for $j=1, 2$.  
Let $\mathcal{S}$ be the sigmoid function $\mathcal{S}(x) = 1/(1+e^{-x})$. 
To jointly predict OU responses $\bm{y}$, we train a neural network which is fed with a pair of UWF fundus images $(\mathcal{X}_1, \mathcal{X}_2)$, and outputs a four-dimensional vector as the predicted values of $\bm{y}$. 
Such a pattern can be characterized by the following nonparametric mean regression model
\begin{align}
    \label{basic model}
\mathbb{E}\{\bm{y}|(\mathcal{X}_1, \mathcal{X}_2)\} = [\bm{\beta}_1^{T} h_1(\mathcal{X}_1), \bm{\beta}_2^{T} h_2(\mathcal{X}_2) ,  \mathcal{S} \{ \bm{\beta}_3^{T} h_1(\mathcal{X}_1)\}, \mathcal{S}\{ \bm{\beta}_4^{T} h_2(\mathcal{X}_2)\}]^{{T}}, 
\end{align}
where $h_j: \mathbb{R}^{224 \times 224 \times 3} \to \mathbb{R}^{d+1}$ denotes an \textit{encoder} that extracts the features of the $j$th UWF fundus image as a ${d+1}$-dimensional representation vector (with the first dimension fixed as $1$ to incorporate the bias term), for $j=1, 2$, and $\bm{\beta}_k \in \mathbb{R}^{d+1} $ denotes the vector of weights (including the bias term) in the regression/classification heads that are connected to the response $y_k$ for $k=1, \ldots, 4$. 
The output dimension $d$ of the encoder $h_j$ is fully determined by the architecture of the neural network.

\subsection{Architecture}
The traditional Vision Transformer \citep[ViT,][]{dosovitskiy2020image} handles a single input image by expressing the representations of an image as a ``class token" vector. 
The regression and classification tasks are then performed by linear learners with the same input of class token.
To better capture the heterogeneity between a pair of UWF fundus images, we introduce a bi-channel architecture to the ViT encoder through residual adapters. 
In the following, we begin with formulating the traditional ViT architecture and then introduce how the proposed bi-channel adapters work on the ViT. 

\noindent{\textbf{Traditional ViT encoder}}. 
In the first step, the ViT splits the UWF fundus image $\mathcal{X} \in \mathbb{R}^{224 \times 224 \times 3}$ to $16 \times 16$ patches. 
The induced 3-channel $(224/16)^2 = 196$ patches are then flattened as a vector $\bm{Z} \in \mathbb{R}^{768}$. 
The flattened vector $\bm{Z}$ is then projected to the \textit{patch embedding layer} $\mathcal{E}_{PE}: \mathbb{R}^{768 \times 196} \to \mathbb{R}^{768 \times 197}$: 
\begin{align}
    \label{patch embedding}
    \mathcal{E}_{PE}(\bm{Z}) = (\bm{\mathcal{T}}_{cls}, \bm{E}\bm{Z}) + \bm{E}_{pos},   
\end{align}
where $\bm{E}, \bm{E}_{pos} \in \mathbb{R}^{768 \times 197}$ denote the weight and positional embedding matrices respectively, and  $\bm{\mathcal{T}}_{cls} \in \mathbb{R}^{768}$ represents the additional \textit{class token} for supervised learning tasks. 
The output of patch embedding layer $\mathcal{E}_{PE}$ is therefore a matrix with $197$ columns of tokens, each of which has length $768$. 

The $197$ tokens are then fed to the \textit{multi-head self-attention layers} $\mathcal{E}_{l}^{(SA)}: \mathbb{R}^{768 \times 197} \to \mathbb{R}^{768 \times 197}$ for $l=1, \ldots, 12$:
\begin{align}
    \label{self attention layer}
    \mathcal{E}_{l}^{(SA)}(\bm{Z}) = \bm{Z} +  \sum_{s=1}^S \bm{W}^{(O)}_{s, l} \left(\bm{W}_{s, l}^{(V)} \bm{Z}\right) \sigma_{soft}\left\{\frac{\left(\bm{W}_{s,l}^{(K)} \bm{Z}\right)^T\left(\Qmat \bm{Z}\right)}{\sqrt{\text{dim}(\Kmat)}}\right\}, 
\end{align}
where $\Qmat, \Kmat, \Vmat \in \mathbb{R}^{64 \times 768}$ and $\Promat \in \mathbb{R}^{769 \times 64}$ denote the weight matrices of value, key, query, and projection respectively, and $\sigma_{soft}$ represents the pointwise Softmax operation. 

The self-attention layer is connected to a \textit{two-layer MLP}  $\mathcal{E}_{l}^{(MLP)}: \mathbb{R}^{768 \times 197} \to \mathbb{R}^{768 \times 197}$:
\begin{align}
    \label{MLP layer}
    \mathcal{E}_{l}^{(MLP)}(\bm{Z}) = \bm{Z}  +  \bm{W}_l^{(2)} \sigma_{G} \left\{ \MLPfir \bm{Z} + \MLPBiasfir \bm{1}_{197}^T \right\} + \MLPBiassec \bm{1}_{197}^{T}, 
\end{align}
where $\MLPfir \in \mathbb{R}^{3072 \times 768}$ and $\MLPsec \in \mathbb{R}^{768 \times 3072}$ denote the weights in the first and second layer of the MLP, respectively, $\MLPBiasfir \in \mathbb{R}^{3072}$ and $\MLPBiassec \in \mathbb{R}^{768}$ denote the bias vectors of the first and second layer of the MLP, respectively, and $\sigma_{G}$ represents the Gaussian Error Linear Units (GeLU) activation: $\sigma_G(x) = x \Phi(x)$. 

After $L$ transformer blocks, we use the class token $\bm{\mathcal{T}}_{cls}$ only as the extracted feature for our multi-response regression. 
Thus, we define the ViT encoder class 
\begin{align*}
    \mathcal{T}_{d, N, S, U, V, L} = \left\{
    \mathcal{E}_L^{(MLP)} \circ \mathcal{E}_L^{(SA)}, \ldots,  \mathcal{E}_1^{(MLP)} \circ \mathcal{E}_1^{(SA)} \circ \mathcal{E}_{PE}
    \right\}(\cdot, 1) \in \mathbb{R}^d, 
\end{align*}
where $d$ denotes the embedding dimension of the encoder (768 in our ViT encoder), $N$ denotes the total number of tokens (197 in our ViT backbone), $S$ denotes the number of self-attention heads, $U$ denotes the head size, $V$ denotes the hidden dimension in the MLP layer, $L$ denotes the stacking number of Transformer blocks, and $(\cdot, 1)$ means we retain the first column only. 
In the standard Vi-base backbone, we have 
$$
(d, N, S, U, V, L) = (768, 197, 12, 64, 3072, 12). 
$$

\begin{figure}[tb]
    \centering
    \includegraphics[scale = .4]{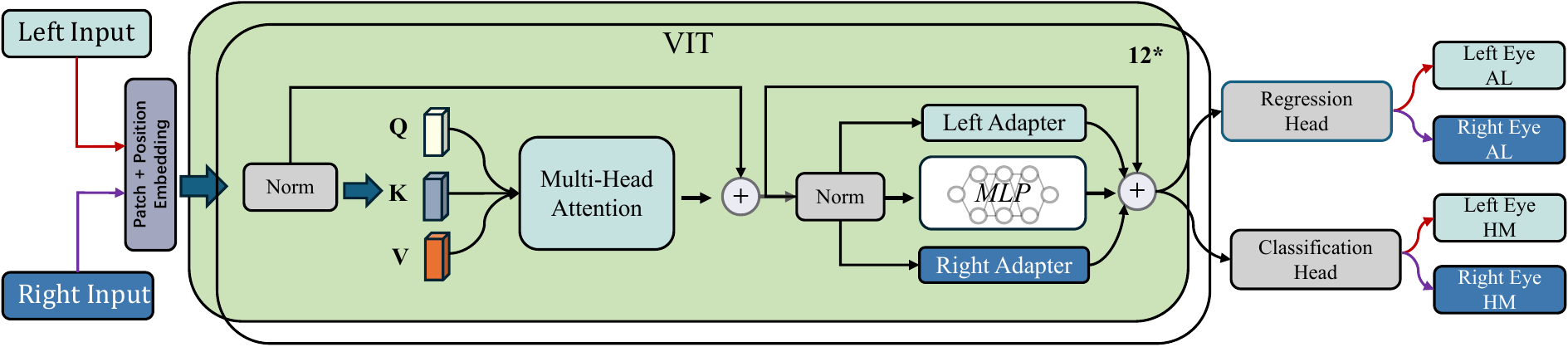}
    \caption{The architecture of the proposed bi-channel ViT.  }
    \label{fig:architecture}
\end{figure}

\noindent{\textbf{Adapter-based bi-channel architecture of ViT.}}
To capture the heterogeneity between OU UWF fundus images, we design a bi-channel architecture on the ViT backbone.  
We present the architecture of the proposed bi-channel ViT  in Figure \ref{fig:architecture}. 
Let $\bm{Z}_{lj}^{SA} \in \mathbb{R}^{768}$ be the output vector of the $l$th  self attention layer $\mathcal{E}_l^{\text{(SA})}$ with the input of image $\mathcal{X}_j$, for $l=1, \ldots, 12$, and $j=1, 2$. 
The self-attention output vector $\bm{Z}_{lj}^{SA}$ is then input to two parallel structures: one is the common  $l$th MLP layers $\mathcal{E}_l^{MLP}$ shared by two eyes, and the other is eye-specific residual adapter $\mathcal{A}_{lj}: \mathbb{R}^{768} \to \mathbb{R}^{768}$ such that
$$
\mathcal{A}_{lj}(\bm{Z}_{lj}) =  \{\text{ReLU}(\bm{V}_{lj1}^\top \bm{Z}_{lj} + \bm{d}_{lj1})\}\bm{V}_{lj2} + \bm{d}_{lj2}, ~j=1, 2, 
$$
where $\bm{V}_{lj1}, \bm{V}_{lj2} \in \mathbb{R}^{768}$, are adapter weights $\bm{d}_{lj1} \in \mathbb{R}$ is the first layer adapter bias, $\bm{d}_{lj2} \in \mathbb{R}^{768}$ is the second layer bias, and $\text{ReLU}$ denotes the ReLU activation. 
The adapter $\mathcal{A}_{lj}$ takes a structure similar to the original ViT MLP layer, but keeps the dimension of vector $\bm{Z}_{lj}$ in all hidden layers instead of dimension increasing. 
We adopt this low-dimensional setting to reduce the model complexity and thus partially remedy the overfitting problem. 
The outputs of the original MLP and the two eye-specific adapters are combined through the following residual link 
\begin{align}
    \label{adapters}
    \mathcal{E}_{lj}^{Ada}(\bm{Z}_{lj}) = \mathcal{E}_l^{MLP}(\bm{Z}_{lj}) + \alpha_j \mathcal{A}_{lj}(\bm{Z}_{lj}), 
\end{align}
where $0<\alpha_j <1$ is the weight imposed on the adapter regarding left and right eyes respectively. 
We allow the adapter weights $\alpha_j$ in \eqref{adapters} to be trainable together with other weights. 
To ensure $\alpha_j \in (0, 1)$, we simply set $\alpha_j = \mathcal{S}(\tilde{\alpha}_j)$, where $\tilde{\alpha}_j \in \mathbb{R}$ has no constraints.

We call $\mathcal{E}_l^{Ada}$ the $l$th adapter block in our bi-channel ViT architecture.  
The adapter block captures the heterogeneity (inter-ocular asymmetry) between two eyes through the residual adapters $\mathcal{A}_{lj}$ for $j=1, 2$, and possesses the inherent dependence (structural similarities)  between the pair-eye UWF fundus images through the shared MLP layer  $\mathcal{M}_l^*$. 
By replacing the original MLP layers to adapter block $\mathcal{E}_l^{Ada}$, we obtain our bi-channel ViT encoder class in our mean regression model \eqref{basic model}.
For $j=1, 2$, 
\begin{align}
    \label{bi-channel encoder}
    h_j:= \mathcal{B}_{d, N, S, U, V, L}^{(j)} = \left\{
    \mathcal{E}_{Lj}^{(Ada)} \circ \mathcal{E}_L^{(SA)}, \ldots,  \mathcal{E}_{Lj}^{(Ada)} \circ \mathcal{E}_1^{(SA)} \circ \mathcal{E}_{PE}
    \right\}(\cdot, 1) \in \mathbb{R}^d. 
\end{align}

\begin{remark}
The adapter-based bi-channel architecture of CeViT differs fundamentally from that of OU-Copula \citep{li2024oucopula}. 
Specifically, we only impose adapters on the MLP layers of the ViT and keep the original multi-head attentional layers frozen. 
The CeViT bi-channel design is motivated by the architecture of the ViT: attention layers model spatial dependencies of image patches that remain anatomically consistent between two eyes, whereas MLP layers perform nonlinear feature transformations which may differ between eyes due to inter-ocular asymmetry. 
Furthermore, to enhance generalization ability, CeViT assigns distinct, trainable weights to each eye, contrasting with the identical and fixed weight in the OU-Copula. 
    
\end{remark}

\subsection{Four-dim mixed copula loss}
In the training procedure, we propose a $4$-dimensional mixed copula loss to capture the conditional dependence structure of the mixed responses $\bm{y}$ given OU image covariates $(\mathcal{X}_1, \mathcal{X}_2)$, as a replacement for the empirical loss that ignores the conditional dependence.
According to \cite{zhong2025cecnn}, it is natural to assume that the OU ALs are Gaussian given the corresponding UWF fundus images.  
Thus, based on model \eqref{basic model}, we model the marginal distributions of $\bm{y}|(\mathcal{X}_1, \mathcal{X}_2)$as 
\begin{eqnarray}
\label{marginal distribution}
    \begin{aligned}
    y_1|\mathcal{X}_1 \sim N(\bm{\beta}_1^T h_1(\mathcal{X}_1), \sigma_1^2), ~~y_3|\mathcal{X}_1 \sim \text{Bernoulli}[\mathcal{S}\{ \bm{\beta}_3^T\ h_1(\mathcal{X}_1)\}],
    \\
y_2|\mathcal{X}_2 \sim N(\bm{\beta}_2^Th_2(\mathcal{X}_2), \sigma_2^2), ~~y_4|\mathcal{X}_2 \sim \text{Bernoulli}[\mathcal{S} \{\bm{\beta}_4^Th_2(\mathcal{X}_2)\}].     \\
\end{aligned}
\end{eqnarray}
With the above marginal conditional distributions, one can specify the joint distribution of $\bm{y}|
(\mathcal{X}_1, \mathcal{X}_2)$ through a Gaussian copula. 

We first present the latent variable expressions for the discrete variables $y_2$ and $y_4$. 
Without ambiguity, let $\mu_{1} = \bm{\beta}_1 ^T h_1(\mathcal{X}_1), \mu_{2}  = \bm{\beta}_2 ^T h_2(\mathcal{X}_2), \mu_{3} :=  \bm{\beta}_3 ^Th_1(\mathcal{X}_1), \mu_{4} :=  \bm{\beta}_4 ^T h_2(\mathcal{X}_2)$. 
Let $v_3, v_4 \sim \mathcal{L}\mathcal{O}$ be two latent variables that marginally follow the standard logistic distribution and are independent to $\mathcal{X}_1$ and $\mathcal{X}_2$. 
Let $X \overset{~~d~~}{=} Y$ be the equality in distribution between random variables $X$ and $Y$. 
We immediately have the following marginal distributions
\begin{align}
    \label{latent variable}
    y_3|\mathcal{X}_1 \overset{~~d~~}{=} I(v_3 + \mu_3 \ge 0), ~ y_4|\mathcal{X}_2 \overset{~~d~~}{=} I(v_4 + \mu_4 \ge 0). 
\end{align}
Let $e_1 = \sigma_1^{-1}(y_1 - \mathbb{E}(y_1)), e_2 = \sigma_2^{-1}(y_2 - \mathbb{E}(y_2))$ be the standardized residuals. 
Let $\bm{u} = [e_1, e_2, \Phi^{-1} \circ \mathcal{S}(v_3), \Phi^{-1} \circ \mathcal{S}(v_4)]^{{T}} := (u_1, u_2, u_3, u_4)^{{T}}$. 
The joint distribution of $\bm{y}|(\mathcal{X}_1, \mathcal{X}_2)$ can be fully expressed by the joint distribution of $\bm{u}$, which is distributed as multivariate normal $\text{MVN}_4(\bm{0}_4, \Gamma)$, where the covariance matrix $\Gamma$ should have a unit diagonal and consequently, can be explicitly expressed as
\begin{align}
\label{Gamma matrix}
\Gamma =  \begin{pmatrix}
1 & \rho_{12} & \rho_{13} & \rho_{14} \\
\rho_{12} & 1 & \rho_{23} & \rho_{24} \\
\rho_{13} & \rho_{23} & 1 & \rho_{34} \\
\rho_{14} & \rho_{24} & \rho_{34} & 1
\end{pmatrix} := \begin{pmatrix}
    \Gamma_{11} & \Gamma_{12}\\
    \Gamma_{21} & \Gamma_{22}
\end{pmatrix}, ~ \Gamma_{ij} \in \mathbb{R}^{2 \times 2}, ~i, j = 1, 2. 
\end{align}

By definition, $\rho_{st} = corr(u_s, u_t)$. 
We call $\mathcal{S}_C = \{\sigma_1, \sigma_2, \rho_{12},\ldots, \rho_{34}\}$ the set of copula parameters hereafter. 
Let $\Phi_2[(\cdot, \cdot); \bm{\mu}, V]$ be the joint CDF of 
$\text{MVN}_2 (\bm{\mu}, V)$. 
Based on the above notation, the log density of $\bm{y}|(\mathcal{X}_1, \mathcal{X}_2)$ is given by the following theorem. 
\begin{theorem}
\label{theo: joint density}
Under mean regression \eqref{basic model} and marginal model \eqref{marginal distribution}, the joint log density of $\bm{y}|(\mathcal{X}_1, \mathcal{X}_2)$ is 
\begin{eqnarray}
    \label{joint log density}
    \begin{aligned}
         l\{\bm{y}|(\mathcal{X}_1, \mathcal{X}_2)\} &= -\frac{1}{2(1-\rho_{12}^2)} \left\{
\frac{(y_1 - \mu_1)^2}{\sigma_1^2}
- \frac{2\rho(y_1 - \mu_1)(y_2 - \mu_2)}{\sigma_1 \sigma_2}
+ \frac{(y_2 - \mu_2)^2}{\sigma_2^2}
\right\}\\
& +\log \big[y_4(1-2y_3)\Phi_{\widetilde{\mu}_1, \widetilde{v}_{11}}(\Phi^{-1} \circ \mathcal{S}(-\mu_3)) + y_3(1-2y_4)\Phi_{\widetilde{\mu}_2, \widetilde{v}_{22}}\{\Phi^{-1} \circ \mathcal{S}(-\mu_4)\} \\
& +(1-2y_3)(1-2y_4)\Phi_2\{(-\mu_3, -\mu_4)^{T}; \widetilde{\bm{\mu}}, \widetilde{V}\} +  y_3y_4 \big] + C,  
    \end{aligned}
\end{eqnarray}
where $\widetilde{\bm{\mu}} = \Gamma_{21}\Gamma_{11}^{-1} (u_1, u_2)^{T} :=  (\widetilde{\mu}_1, \widetilde{\mu}_2)^{T}$, $\widetilde{V} = \Gamma_{22} - \Gamma_{21} \Gamma_{11}^{-1} \Gamma_{12}$, 
and $C$ is the normalization constant. 
\end{theorem}

\begin{remark}
One can show the equivalence between \eqref{joint log density} and the general likelihood form \citep{song2009joint}, which needs to compute general multivariate Gaussian integrals with various lower and upper bounds. 
Instead, \eqref{joint log density} transforms general multivariate Gaussian integrals into multivariate/univariate CDFs by employing the inclusion-exclusion principle, which are available in PyTorch through the Torch-MvNorm library \citep{marmin2015differentiating}. 
Similarly, one can extend Theorem \ref{theo: joint density} to general $p$-dimensional Gaussian copula models with $p_0$ regression tasks and $(p - p_0)$ classification tasks in PyTorch; refer to Corollary A.1 in the Supplement. 
\end{remark}

Let $n$ be the data size. 
Let $\mathcal{D} = \{\bm{y}_i, \mathcal{X}_{i1}, \mathcal{X}_{i2}\}_{i=1}^n$ be the dataset of training labels and training images. 
Let $\bm{W}$ be the weights of CeViT. 
Based on Theorem \ref{theo: joint density}, the copula loss function is given by the negative log likelihood
\begin{align}
    \label{copula loss}
\mathcal{L}\{\bm{W}|(\mathcal{D}, \mathcal{S}_C)\} = \sum_{i=1}^n -l\{\bm{y}_i|(\mathcal{X}_{i1}, \mathcal{X}_{i2})\}. 
\end{align}

\subsection{Fast Monte Carlo Expectation-Maximization (fMCEM)}
The proposed copula loss \eqref{copula loss} is based on the known copula parameters $\mathcal{S}_C$. 
The key problem that we need to address before employing the copula loss is the estimation of $\mathcal{S}_C$. 
On continuous margins, estimation of $\sigma_1$  and $\sigma_2$ is straightforward based on empirical estimators 
$$\hat{\sigma}_1 = \text{SD}(u_1),  \hat{\sigma}_2 = \text{SD}(u_2), 
$$
where $\text{SD}$  represents the sample standard deviation. 
However, estimation of the parameters $(\rho_{13}, \rho_{14}, \rho_{23}, \rho_{24}, \rho_{34})$ is difficult, since the latent variables on discrete margins, i.e. $u_3$ and $u_4$, are unobservable. 
Fortunately, the information of the two latent variables is no longer a black-box once we observe the standardized residuals $u_1, u_2$, the discrete responses $y_2$ and $y_4$, and outputs of the neural network on the classification heads $\mu_3$ and $\mu_4$.

\noindent{\textbf{Posterior of latent variables.}}
Based on latent variable expression \eqref{latent variable}, we have, marginally, 
\begin{align*}
    u_3|y_3 \sim y_3TN(0, 1, -\Phi^{-1}\circ \mathcal{S}(\mu_3), +\infty) &+ (1-y_3)TN(0, 1, -\infty,  -\Phi^{-1}\circ \mathcal{S}(\mu_3)) \\
u_4|y_4 \sim y_4TN(0, 1, -\Phi^{-1}\circ \mathcal{S}(\mu_4), +\infty) &+ (1-y_4)TN(0, 1, -\infty,  -\Phi^{-1}\circ \mathcal{S}(\mu_4)),  
\end{align*}
where $TN(\mu, \sigma^2, a, b)$ denotes the truncated normal distribution of $N(\mu, \sigma^2)$ with lower and upper bounds of $a$ and $b$ respectively.

Let $\mathcal{O} = \{u_1, u_2, y_2, y_4, \mu_3, \mu_4\}$ be the observed variables. 
Given $\Gamma$, the posterior distribution of $(u_3, u_4)$ is given by 
\begin{align}
    \label{latent posterior}
  (u_3, u_4)|(\Gamma, \mathcal{O} )\sim \text{TMvN}(\tilde{\mu}, \tilde{V}; \bm{a}_{\mathcal{O}}, \bm{b}_{\mathcal{O}}), 
\end{align}
where $\tilde{\mu}$ and $\tilde{V}$ take the same form in Theorem \ref{theo: joint density}, and $\text{TMvN}(\tilde{\mu}, \tilde{V}; \bm{a}_{\mathcal{O}}, \bm{b}_{\mathcal{O}})$ denotes the truncated multivariate normal distribution with mean $\tilde{\mu}$ and covariance $\tilde{V}$, and the lower \& upper bounds $\bm{a}_{\mathcal{O}}$ and $\bm{b}_{\mathcal{O}}$ are given by 
\begin{align*}
    \bm{a}_{\mathcal{O}} =\{-y_3\Phi^{-1}\circ \mathcal{S}(\mu_3) - (1-y_3)\infty, ~ -y_4\Phi^{-1}\circ \mathcal{S}(\mu_4) - (1-y_4)\infty\}, \\
    \bm{b }_{\mathcal{O}} = \{-(1-y_3)\Phi^{-1}\circ \mathcal{S}(\mu_3) + y_3\infty, ~ -(1-y_4)\Phi^{-1}\circ \mathcal{S}(\mu_4) + y_4\infty\}. 
\end{align*}
Denote the multivariate normal log-likelihood of the observed-latent variables  with covariance  matrix ${\Gamma}$ by $\log \text{L}_{{\Gamma}}$.
Denote an estimator of $\Gamma$ by $\hat{\Gamma}$, where the four $2\times 2$ sub-matrices are denoted by $\hat{\Gamma}_{ij}$, $i, j = 1, 2$, respectively. 
An EM algorithm for $\Gamma$ estimation is conducted by iteratively maximizing the expected log-likelihood $\mathbb{E}_{(u_{3i}, u_{4i})|(\hat{\Gamma}, \mathcal{O}_i)} \log \text{L}_{\hat{\Gamma}}$, where the expectation is computed under the posterior distribution \eqref{latent posterior}. 
Since the expected log-likelihood  does not have a closed form, we consider an Monte Carlo EM (MCEM) \citep{wei1990monte} algorithm to sample the expected log-likelihood instead. 
We present the MCEM procedure in pseudo Algorithm \ref{alg: MCEM}.

\begin{algorithm*}[!htb]\footnotesize
\caption{Pseudo algorithm of MCEM for $\Gamma$ estimation}\label{alg: MCEM}
\begin{algorithmic}[1]
\Require Observed variables $\mathcal{O}$; Monte Carlo replication $M$; iteration upper bound $T$; initial estimator $\hat{\Gamma}^{(0)}$; data size $n$. 
\Ensure Estimator $\hat{\Gamma}$. 
\For{$t \gets 1$ to $T$}
\State  \textbf{E-step}. For $i =1, \ldots, n$, draw $M$ samples $(u_{3i}^{(m)}, u_{4i}^{(m)})$ for $m=1, \ldots, M$ from posterior $(u_{3i}, u_{4i})|(\hat{\Gamma}^{(t-1)}, \mathcal{O}_i)$ based on \eqref{latent posterior}. Let $q(\Gamma, \mathcal{O}_i) := \mathbb{E}_{(u_{3i}, u_{4i})|(\hat{\Gamma}^{(t-1)}, \mathcal{O}_i)} \log \text{L}_{\Gamma} = M^{-1} \sum_{m=1}^M \log \text{L}_{\Gamma} \left(u_{3i}^{(m)}, u_{4i}^{(m)}, \mathcal{O}_i\right) $. 
\State \textbf{M-step}. Define $(\Gamma, \mathcal{O}) :=  \sum_{i=1}^n q(\Gamma, \mathcal{O}_i)$. 
$\hat{\Gamma}^{(t)} \gets \arg \max \limits_{\Gamma} q(\Gamma, \mathcal{O})$. 
\EndFor
\end{algorithmic}
\end{algorithm*}

Though the MCEM Algorithm \ref{alg: MCEM} is straightforward, it is numerically difficult to implement in Python due to two challenges: i) sampling from truncated multivariate normal distribution in the E-step is not efficient and unstable in Python; ii) the solution of maximizing the expected log-likelihood $q(\Gamma, \mathcal{O})$ in the M-step is non-trivial. 
Therefore, in the next step, we attempt to simplify and speed up the MCEM Algorithm \ref{alg: MCEM}.

\noindent{\textbf{Fast MCEM algorithm}}. 
For $i=1, \ldots, n$, with a pair of samples $\left(u_{3i}^{(m)}, u_{4i}^{(m)}\right)$, the log-likelihood contribution takes the sandwich form (up to a constant): 
$
\log L_{\Gamma}\left(u_{3i}^{(m)}, u_{4i}^{(m)}, \mathcal{O}_i\right) = \log(|\Gamma|) - \frac{1}{2} \bm{z}_{mi}^{\mathsf{T}} \Gamma^{-1} \bm{z}_{mi}
$, 
where $\bm{z}_{mi}  = \left(u_{1i}, u_{2i}, u_{3i}^{(m)}, u_{4i}^{(m)}\right)^{\mathsf{T}}$. 
Let $\bar{\bm{z}}_i$ and $\Sigma_{\bm{z}_i}$ be the sample mean vector and sample covariance matrix of samples $\bm{z}_{mi}$, respectively. 
By some algebra, we have 
\begin{eqnarray*}
    \begin{aligned}
        q(\Gamma, \mathcal{O}_i) & = M^{-1} \sum_{m=1}^M \log \text{L}_{\Gamma} \left(u_{3i}^{(m)}, u_{4i}^{(m)}, \mathcal{O}_i\right) 
         = \log(|\Gamma|) - \frac{1}{2}\text{tr}\{\Gamma^{-1}(M^{-1}\bm{z}_{mi} \bm{z}_{mi}^{\mathsf{T}} )\}\\ 
        &= \log(|\Gamma|) - \frac{1}{2}\text{tr}\{\Gamma^{-1}(\Sigma_{\bm{z}_i} +  \Bar{\bm{z}}_i \bar{\bm{z}}_i^{\mathsf{T}})\} = \log(|\Gamma|) - \frac{1}{2}\text{tr}(\Gamma^{-1}\Sigma_{\bm{z}_i}) - \frac{1}{2}\bar{\bm{z}}_i^{\mathsf{T}} \Gamma^{-1} \bar{\bm{z}}_i. 
    \end{aligned}
\end{eqnarray*}

Since $u_{i1}, u_{i2}$ are fixed, the sample covariance reduces to 
$$
\Sigma_{\bm{z}_i}=  \begin{pmatrix}
0 & 0 & 0 & 0 \\
0 & 0 & 0 & 0 \\
0 & 0 & s_{33i} & s_{34i} \\
0 &0 & s_{34i} & s_{44i} 
\end{pmatrix} :=  \begin{pmatrix}
    0 & 0\\
        0 & \Sigma_{22i}
\end{pmatrix}. 
$$
Since $(u_{3i}, u_{4i})$ are sampled from posterior \eqref{latent posterior}, an intuitive approximation of $\Sigma_{22i}$ is the conditional covariance matrix $\tilde{V}$ in Theorem \ref{theo: joint density}. 
Under this approximation, $\text{tr}(\Gamma^{-1}\Sigma_{\bm{z}_i}) = 2$ for all $i=1, \ldots, n$, which is invariant and unrelated to $\Gamma$. 
Meanwhile, as $M \to \infty$, $\bar{\bm{z}}_i \to (u_{i1}, u_{i2}, \mathbb{E}_{u_3|(\Gamma, \mathcal{O}_i)}, \mathbb{E}_{u_4|(\Gamma, \mathcal{O}_i)})$. 
Thus, we obtain the following approximation 
\begin{align}
\label{approximate energy}
    q(\Gamma, \mathcal{O}) = \sum_{i=1}^n q(\Gamma, \mathcal{O}_i) \approx n \log(|\Gamma|) - \frac{1}{2}\sum_{i=1}^n \bar{\bm{z}}_i^{\mathsf{T}} \Gamma^{-1} \bar{\bm{z}}_i +C,  
\end{align}
where $C$ is some constant unrelated to $\Gamma$. 
It is trivial that the approximate expected log-likelihood \eqref{approximate energy} is maximized by the sample correlation matrix of $\bm{\bm{z}}_i$. 
Thus, we have the fast Monte Carlo Expectation Maximization (fMCEM) Algorithm \ref{alg: FMCEM}.
The estimation of copula parameters $\mathcal{S}_C$ is accordingly given by $\{\hat{\sigma}_1, \hat{\sigma}_2, \hat{\Gamma}_{\text{off}} \}$, where $\hat{\Gamma}_{\text{off}}$ denotes the off-diagonal elements of $\hat{\Gamma}$. 

The proposed fMCEM Algorithm is computationally efficient since its E-step computes $2n$ truncated normal expectations, which have closed forms. 
In contrast, the extended rank-likelihood MCMC approach \citep{hoff2007extending} requires sampling $2n$ truncated normal samples in each Markov state, leading to much heavier computational burden. 
Furthermore, empirically, the FMCEM algorithm can converge in several steps, while the MCMC procedure needs to draw a long Markov chain. 

\begin{algorithm*}[!htb]\footnotesize
\caption{fMCEM algorithm for $\Gamma$ estimation}\label{alg: FMCEM}
\begin{algorithmic}[1]
\Require Observed variables $\mathcal{O}$; iteration upper bound $T$; initial estimator $\hat{\Gamma}^{(0)}$; data size $n$. 
\Ensure Estimator $\hat{\Gamma}$. 
\For{$t \gets 1$ to $T$}
\State  \textbf{E-step}. For $i =1, \ldots, n$, $\bm{z}_i \gets \left(u_{i1}, u_{i2}, \mathbb{E}_{u_3|(\hat{\Gamma}^{(t-1)}, \mathcal{O}_i)}, \mathbb{E}_{u_4|(\hat{\Gamma}^{(t-1)}, \mathcal{O}_i)}\right)$. 
\State \textbf{M-step}. Let $\bm{Z} = (\bm{z}_1,\ldots, \bm{z}_n)$. 
Update $\hat{\Gamma}^{(t)} \gets \text{cov2cor} (n^{-1}\bm{Z}\bm{Z}^{\mathsf{T}} )$, where $\text{cov2cor}$ denotes the operator that transfers a covariance matrix to a correlation matrix. 
\EndFor
\end{algorithmic}
\end{algorithm*}

\begin{remark}
 We admit that generally $\Sigma_{22i} \not = \tilde{V}$ since the sample covariance of a multivariate truncated normal distribution $\text{TMvN}(\tilde{\mu}, \tilde{V}; \bm{a}_{\mathcal{O}}, \bm{b}_{\mathcal{O}})$ will not converge to $\tilde{V}$. 
As a result, given consistent estimates of $\bm{a}_{\mathcal{O}}$ and $\bm{b}_{\mathcal{O}}$, the FMCEM tends to underestimate $\rho_{34}$; refer to the Supplement for details. 
Nevertheless, we argue that such an underestimation is beneficial in enhancing ViT's predictive capability under the overfitting nature of deep learning models, while the consistent extended rank-likelihood MCMC approach may encounter numerical instability; refer to Section \ref{sec: overfitting} for detailed discussions. 
\end{remark}

\section{Feasible CeViT, fine tuning, and prediction}
\label{sec: feasible}
This section introduces the training framework, fine tuning details, rationale of the underestimated fMCEM algorithm, and a new prediction scheme based the aforementioned ViT architecture and the proposed copula loss.

\subsection{Feasible CeViT framework}
We call our training framework feasible CeViT as a salute to the feasible generalized least squares estimator \citep{avery1977error}, which estimates the correlation matrix based on the residuals for ordinary least square estimation. 
In essence, our feasible CeViT incorporates a warm-up ViT to obtain the observed variables $\mathcal{O}$, an estimation module employing the fMCEM algorithm to estimate copula parameters $\mathcal{S}_C$, and a final training procedure. 
Let $\mathcal{L}_0$ be the following empirical loss for the weights $\bm{W}$ in our bi-channel ViT
\begin{eqnarray}
        \label{empirical loss}
\begin{aligned}
  \mathcal{L}_0(\bm{W} |\mathcal{D}) =  \sum_{i=1}^n \{(\hat{y}_{i1}-y_{i1})^2 +   (\hat{y}_{i2}-y_{i2})^2 + 
\text{BCE}(\hat{\mu}_{i3}, y_{i3}) + \text{BCE}(\hat{\mu}_{i4}, y_{i4})\}, 
\end{aligned}
\end{eqnarray}
where $\hat{y}_{i1}, \hat{y}_{i3}$ are the outputs of the regression heads, $\hat{\mu}_{i3}, \hat{\mu}_{i4}$ are the outputs of the classification heads, for left and right eyes respectively, and $\text{BCE}(\mu, y) = y\log(\mu) + (1-y)\log(1-\mu)$ denotes the binary cross entropy loss. 
Then the warm-up ViT is obtained by optimizing the empirical loss \eqref{empirical loss}. 
Let $(\hat{y}_{i10}, \hat{y}_{i20})$ be the outputs of the regression heads of the warm-up ViT. 
The estimation module is accordingly given by setting $\mathcal{O}_i = \{y_{i1} - \hat{y}_{i1}, y_{i2} - \hat{y}_{i2}, y_{i1}, y_{i2}, \hat{\mu}_{i3}, \hat{\mu}_{i4}\}$ for $i=1, \ldots, n$ in the fMCEM  algorithm. 
With the estimated copula parameters $\mathcal{S}_C$, the final training procedure is completed by optimizing the copula loss \eqref{copula loss}. 
We summarize the whole feasible CeViT framework in Algorithm \ref{alg: CeViT}. \\

\begin{algorithm*}[!htb]\footnotesize
\caption{Feasible CeViT algorithm}\label{alg: CeViT}
\begin{algorithmic}[1]
\Require Training data $\mathcal{D}$; data size $n$. 
\Ensure Fitting CeViT with weights $\hat{\bm{W}}$ trained under the proposed copula loss \eqref{copula loss}.  
\State Fit the warm-up ViT model by optimizing empirical loss \eqref{empirical loss}: $\hat{\bm{W}}_0 = \arg \min \limits_{\bm{W}} \mathcal{L}_0 (\bm{W} | \mathcal{D})$ on GPU. 
\State  Obtain empirical estimator of copula parameters $\hat{\mathcal{S}}_C$ by FMCEM Algorithm \ref{alg: FMCEM} on CPU. 
\State Fit the CeViT model by optimizing copula loss \eqref{copula loss}: $\hat{\bm{W}} = \arg \min \limits_{\bm{W}} \mathcal{L} \{\bm{W} | (\mathcal{D}, \hat{\mathcal{S}_C})\} $ on GPU. 
\end{algorithmic}
\end{algorithm*}

\subsection{Fine-tuning with LoRA}
In CeCNN and OU-Copula, the overfitting problem is circumvented by reducing the model complexity. 
However, a simplified AI model generally encounter lower generalizability in predictions \citep{zhai2022scaling}.  
To avoid the lose of generalizability, CeViT preserves the original ViT model size and mitigates the overfitting problem by fine tuning the pretrained ViT encoder to reduce the number of \textit{trainable parameters}. 
Let $\bm{W}_l^{(Q)} = ({\bm{W}_{1, l}^{(Q)}}^T, \ldots, {\bm{W}_{12, l}^{(Q)}}^T) \in \mathbb{R}^{768 \times 768}$ be the value weight matrix of the $l$th layer of ViT encoder, for $l=1. \ldots, 12$. 
The Low-Rank Adaptation (LoRA) introduces a low rank structure on the trainable  weight matrices
$$
\bm{W}_l^{(Q)} = \bm{W}_{l0}^{(Q)} + A_l^{(Q)}B_l^{(Q)}, ~ l=1, \ldots, 12, 
$$
where $\bm{W}_{l0}^{(Q)} \in \mathbb{R}^{768 \times 768}$ is the weight matrix pretrained on the large ImageNet dataset \citep{deng2009imagenet}, and $A_l^{(Q)} \in \mathbb{R}^{768 \times r}, B_l^{(Q)} \in \mathbb{R}^{r \times 768}$ are two rank-$r$ matrices of trainable weights. 
In this paper we set $r=8$ as default. 
During the fine tuning procedure, we freeze the pretrained weight matrix $\bm{W}_{l0}^{(Q)}$ and only update the trainable low-rank matrices $A_l^{(Q)} $ and $B_l ^{(Q)}$ in each iteration, with $A_l^{(Q)}$ and $B_l^{(Q)}$ initialized from Gaussian random weights and zeros respectively. 
Similar LoRA structures are also imposed for fine tuning  weight matrices $\bm{W}_l^{(K)}, \bm{W}_l^{(V)}, \bm{W}_l^{(O)}, \bm{W}_l^{(1)}, \bm{W}_l^{(2)}$ in all multi-head self-attention layers and MLP layers. 
As a result, under LoRA, the trainable parameter size of CeViT is around 1.2 million, which not only plummets from that of the original ViT (85 million), but is also smaller than the size of the popular CNN model ResNet18 \citep{he2016deep} backbone (11.5 million).

In terms of weight matrices $\bm{V}_{lj1}$ in residual adapters for $l, j =1, 2$ and weights $\bm{\beta}_k$  in regression/classification heads for $k=1, \ldots, 4$, we maintain their full-parameter training without LoRA. 
In this sense, our model training can be understood as a ``dual-adapter" scheme, since residual adapters and low-rank adapters are trained in different ways.

\subsection{Prediction scheme based on Bayes classifier}
The CeViT model trained on the training dataset with copula loss \eqref{copula loss} can be applied to predict OU HM and AL for future individuals. 
The prediction of OU AL is straightforward: we use the outputs of the regression heads as the predicted values of OU AL. 
Denote the marginal estimated probability of OU HM by $Pr\{\hat{y}_j = 1\} := \hat{p}_j$, for $j=3, 4$. 
For OU HM classification, we do NOT use the naive classification rule $\hat{y}_j = I(\hat{p}_j > 0.5)$.
Rather, we consider OU HM classification as a four-class classification with the label set $S = \{(1,1), (0, 1), (1, 0), (0,0)\}$. 
Then based on \eqref{latent variable}, we have \begin{align*}
    &\hat{p}_{(1, 1)} = \Phi_{2}\{\Phi^{-1} (1-\hat{p}_3), \Phi^{-1} (1-\hat{p}_4); \bm{0}_2, \hat{\Gamma}_{22}\},~ \hat{p}_{(0, 1)} = \hat{p}_4 - \hat{p}_{(1, 1)}, \\
    &\hat{p}_{(1, 0)} = \hat{p}_3 - \hat{p}_{(1, 1)},\quad  \hat{p}_{(0, 0)} = 1 + \hat{p}_{(1, 1)} - \hat{p}_{(0, 1)} - \hat{p}_{(1, 0)}. 
\end{align*}
The corresponding Bayes classifier is then given by 
$$
(\hat{y}_3, \hat{y}_4) =  \mathcal{C}(\mathcal{X}_1, \mathcal{X}_2)= \arg \max \limits_{s \in S} \hat{p}_s. 
$$

\section{Stronger covariance phenomenon and the disturbance}
\label{sec: overfitting}
Under the CeViT architecture and the proposed copula loss, note that LoRA only mitigates overfitting from two aspects: guaranteeing stable training on limited training data and retaining sufficient generalizability on test sets or external data. 
Nonetheless, overfitting mitigation does NOT mean no overfitting at all. 
Instead, in the multitask learning domain, the overparameterized model may face a special overfitting problem, which we call the ``\textit{stronger covariance phenomenon}" hereafter.

Recall that under the mean regression model \eqref{basic model}, the conditional mean $\mathbb{E} (y_k |\mathcal{X}_j)$ are fully determined by $\mu_k := \bm{\beta}^T h_{j}(\mathcal{X}_j)$, for $k=1, \ldots, 4$ and $j =1, 2$. 
Denote their Gaussian scores as 
$$
q_1=\mu_1, q_2 = \mu_2, q_3 = \Phi^{-1}\circ \mathcal{S} (\mu_3), q_4 = \Phi^{-1}\circ \mathcal{S} (\mu_4)
$$
or equivalently, 
\begin{align*}
    q_1 = \mathbb{E}(y_1|\mathcal{X}_1), q_2 = \mathbb{E}(y_2|\mathcal{X}_2), 
    q_3 = \Phi^{-1}\circ \mathcal{S}\{\mathbb{E}(y_3|\mathcal{X}_1)\}, q_4 = \Phi^{-1}\circ \mathcal{S}\{\mathbb{E}(y_4|\mathcal{X}_2)\}.  
\end{align*}
Denote their corresponding estimates output by a deep neural network as $\hat{q}_k$ respectively for $k=1, \ldots, 4$. 
We say that the stronger covariance phenomenon occurs on pair tasks $y_{k_1}$ and $y_{k_2}$ if 
\begin{align}
 |\text{Cov}(\hat{q}_{k_1}, \hat{q}_{k_2})| > |\text{Cov}(q_{k_1}, q_{k_2})|, ~ k_1 \not = k_2, ~ k_1, k_2 =1, \ldots, 4. 
\end{align}
The stronger covariance phenomenon indicates that the estimated conditional means (or conditional mean Gaussian scores) have stronger covariance than that of the ground truth (in absolute value).

The stronger covariance phenomenon may be substantiated by the random feature modeling of benign overfitting \citep{li2023benign}, where the bias term in a neural network is treated as an additional noise imposed to the true features. 
Recall that in the original ViT encoder, the $L$th MLP layer \eqref{MLP layer} has a specific bias term $\bm{b}_L^{(2)}$ that is shared by both eyes. 
In this sense, the shared bias term  $\bm{b}_L^{(2)}$ acts as if an identical noise imposed simultaneously to the true features of images $\mathcal{X}_1$ and $\mathcal{X}_2$. 
As a result, the shared bias term naturally yields additional covariance between $\hat{q}_{k_1}$ and $\hat{q}_{k_2}$ when $\bm{\beta}_{k_1}$ and $\bm{\beta}_{k_2}$ are NOT orthogonal.   
Although the stronger covariance phenomenon has never been reported in the existing literature, we note that the stronger covariance phenomenon can commonly appear in  general multitask learning practice, since the mainstream multitask learning models have shared features and shared bias term across multiple tasks \citep{zhao2018modulation, liu2019end}. 

The stronger covariance phenomenon has substantial disturbance to estimation of copula parameters: 
overestimation on classification tasks and underestimation on regression tasks.
The former is the fundamental reason for the numerical instability of the extended rank-likelihood MCMC approach \citep{hoff2007extending}. 
On the contrary, the proposed fMCEM algorithm owns an underestimation nature, which remedies the overestimation handicap on classification tasks and thus, ensures stable computation. 

\noindent{\textbf{Overestimation  on classification tasks.}}
Recall that $\rho_{34}= corr(u_3, u_4)$ is defined as the correlation coefficient between latent variables $u_3$ and $u_4$.
Given observed labels $y_3$ and $y_4$, the posterior of $(u_3, u_4)$ follows a truncated multivariate normal distribution \eqref{latent posterior}, where the truncation bounds are given by $\hat{q}_3$ and $\hat{q}_4$ respectively. 
In our application setting, the following proposition states that the overfitting problem of the stronger covariance phenomenon leads to the numerical instability of the extended rank-likelihood MCMC approaches. 

\begin{proposition}[Near singularity of MCMC estimates]
\label{prop: overestimation}
    Let $\tilde{\rho}_{34}$ be a plug-in consistent estimator of $\rho_{34}$ in the sense that $\hat{\rho}_{34}$ is consistent to $\rho_{34}$ given correctly specified truncation bounds $\bm{a}_{\mathcal{O}}$ and $\bm{b}_{\mathcal{O}}$ in the posterior distribution \eqref{latent posterior}. 
    Suppose that $\mathbb{E}Pr\{y_3 = y_4 |(\mathcal{X}_1, \mathcal{X}_2)\} > \mathbb{E}Pr\{y_3 \not= y_4 |(\mathcal{X}_1, \mathcal{X}_2)\}$ and $\text{sgn}(\rho_{34}) = \text{sgn}\{\text{Cov}(q_3, q_4)\} = \text{sgn}\{\text{Cov}(\hat{q}_3, \hat{q}_4)\} $. 
    Then if the stronger covariance phenomenon occurs on $y_3$ and $y_4$, we have $|\mathbb{E}(\tilde{\rho}_{34})| > |\rho_{34}|$. 
    If $|\text{Cov}(\hat{q}_3, \hat{q}_4)|/|\text{Cov}(q_3, q_4)| \to \infty$, $|\mathbb{E}(\rho_{34})| \to 1$. 
\end{proposition}

The two conditions on label concordance and concordant correlation naturally hold in our application setting. 
Note that the posterior mean of the extended rank-likelihood MCMC \citep{hoff2007extending} is a plug-in consistent estimator based on the posterior consistency \citep{murray2013bayesian}. 
The insight of Proposition \ref{prop: overestimation} directly explains our empirical observation that the extended rank-likelihood MCMC approach encounters numerical instability: 
the overestimated $\hat{\rho}_{34}$ approaches $1$, making the induced correlation matrix $\Gamma$ nearly singular. 
Nearly singular $\Gamma$ is dangerous to optimizing the copula loss, since it can easily lead to infinite gradients during backpropagation.

Fortunately, the inconsistent fMCEM algorithm naturally balances the overestimation under the stronger covariance phenomenon. 
The following proposition guarantees the numerical stability of the fMCEM algorithm. 

\begin{proposition}[Numerical stability of fMCEM estimates]
    \label{prop: no over 1}
Let $\hat{\rho}_{34}$ be the estimator of $\rho_{34}$ given by the fMCEM algorithm \ref{alg: FMCEM}. 
Suppose that $\text{sgn}(\rho_{34}) = \text{sgn}\{\text{Cov}(q_3, q_4)\} = \text{sgn}\{\text{Cov}(\hat{q}_3, \hat{q}_4)\} > 0$. 
Then if the stronger covariance phenomenon occurs on $y_3$ and $y_4$, we have $|\mathbb{E}(\hat{\rho}_{34})|< \bar{\rho} < 1$, where $\bar{\rho}$ is some constant depending on $\rho_{34}$ and $\text{Cov}(\hat{q}_3, \hat{q}_4)$. 
\end{proposition}

Proposition \ref{prop: no over 1} states that no matter how large the ratio $|\text{Cov}(\hat{q}_3, \hat{q}_4)|/|\text{Cov}(q_3, q_4)|$ is, the estimated correlation matrix $\Gamma$ will never be singular. 
Despite the sacrifice of the consistency of estimating $\Gamma$, the proposed fMCEM algorithm guarantees the numerical robustness against the stronger covariance phenomenon. 
Therefore, the fMCEM algorithm is widely applicable to general multitask learning problems. 

We note that the overestimation nature of $\rho_{34}$ should be avoided in practice, since overly estimated $\hat{\rho}_{34}$ may lead to numerically singular estimates of $\Gamma$. 
Note that the estimator given by the extended rank-likelihood MCMC approach is also plug-in consistent \citep{murray2013bayesian}. 
This theoretical insight explains our empirical observation that the extended rank-likelihood MCMC approach exhibits numerical instability in our numerical studies.
As a remedy, the underestimated fMCEM algorithm naturally counterbalances the overestimation tendency induced by the stronger covariance phenomenon.
Such a bias cancellation ensures the numerical stability and reliable convergence of the fMCEM algorithm.

\noindent{\textbf{Underestimation  on regression tasks.}} On the other hand, on regression tasks, the stronger covariance phenomenon yields underestimation of $\rho_{12}$, revealed by the following proposition. 

\begin{proposition}[Underestimation on regression tasks]
\label{prop: underestimation regression}
Let $\hat{\rho}_{12}$ be a plug-in consistent estimator of $\rho_{12}$ given that $q_1$ and $q_2$ are consistently estimated.
Suppose that $\text{sgn}(\rho_{12}) = \text{sgn}\{\text{Cov}(q_1, q_2)\} = \text{sgn}\{\text{Cov}(\hat{q}_1, \hat{q}_2)\}$. 
Then if the stronger covariance phenomenon occurs on $y_1$ and $y_2$,
we have $|\mathbb{E}(\hat{\rho}_{12})| < |\rho_{12}|$. 
\end{proposition}

Note that in the estimation of $\rho_{12}$, both the fMCEM algorithm and the extended rank-likelihood MCMC converge to the Pearson correlation coefficient residuals $e_1$ and $e_2$, which is a plug-in consistent estimator of $\rho_{12}$. 
Therefore, the underestimation holds for the two methods.

\begin{remark}[Interpretation of conditional dependence]
  Underestimation of $\rho_{12}$ is not a real problem for regression task since the inherent covariance $\text{Cov}(y_1, y_2)$ still holds. 
  Based on the law of total covariance \begin{eqnarray*}
    \begin{aligned}
        \text{Cov}(y_1, y_2) &= \mathbb{E\text{Cov}}\{(y_1, y_2)|(\mathcal{X}_1, \mathcal{X}_2))\} + \text{Cov}\{\mathbb{E}(y_1|\mathcal{X}_1),\mathbb{E}(y_2|\mathcal{X}_2)\} \\
    &=\rho_{12}\sigma_1\sigma_2 + \text{Cov}(q_1, q_2). 
    \end{aligned}
\end{eqnarray*}
In regression tasks, once overestimated $\text{Cov}(q_1, q_2)$ is overly overestimated, $\rho_{12}$ is underestimated to keep $\text{Cov}(y_1, y_2)$ fixed. 
However, on classification tasks, $\text{Cov}(y_3, y_4)$ is not a linear function of the latent correlation coefficient $\rho_{34}$. 
In precise, we have 
$$
\text{Cov}\{(y_3, y_4)|(\mathcal{X}_1, \mathcal{X}_2)\} = \Phi_2\{(q_3, q_4); \bm{0}_2, \Gamma_{22}\} - \Phi(q_3) \Phi(q_4).  
$$
One can show that for fixed $\rho_{34}$, $\text{Cov}\{(y_3, y_4)|(\mathcal{X}_1, \mathcal{X}_2)\}$ is an increasing function of $\text{Cov}(q_3, q_4)$; or conversely, for fixed $\text{Cov}(q_3, q_4)$, it increases with $\rho_{34}$. 
Consequently, overestimation of $\text{Cov}(\hat{q}_3, \hat{q}_4)$ propagates to  $\hat{\rho}_{34}$. 
\end{remark}

\section{Simulation}
\label{sec:simulation}
We conduct simulations to illustrate the proposed CeViT method. 
We use a synthetic UWF fundus image example to validate the efficacy of the proposed copula loss and verify the disturbance of the stronger covariance phenomenon. 
An additional toy example evaluating the computational efficiency of the fMCEM algorithm is deferred to the Supplement. 

\noindent{\textbf{Data generation.}}
We generate a pair of synthetic OU UWF image  $\mathcal{X}_1$ and $\mathcal{X}_2$ first as the covariates. 
Based on paired image covariates, we generate continuous responses (AL) $y_{1}, y_{2} \in \mathbb{R}$ and binary responses (myopia status) $y_{3}, y_{4} \in \{0, 1\}$. 
Let $g_1$ and $g_2$ be the nonlinear regression functions for the continuous and discrete responses respectively. 
We first generate the model error vector $\bm{\epsilon} = (\epsilon_1, \epsilon_2, \epsilon_3, \epsilon_4)^T \sim N(\bm{0}_4, \Sigma_\epsilon)$ with the covariance matrix:
$$ \Sigma_\epsilon = \begin{pmatrix} 1 & 0.720 & 0.294 & 0.213 \\ 0.720 & 1 & 0.205 & 0.336 \\ 0.294 & 0.205 & 1 & 0.569 \\ 0.213 & 0.336 & 0.569 & 1 \end{pmatrix}. $$
Then we generate the four responses through the following multiresponse regression model:
\begin{align*}
    &y_{1} = g_1(\mathcal{X}_1) + \epsilon_1, \quad y_{2} = g_{1}(\mathcal{X}_2) + \epsilon_2, \\
    &y_{3} = I\{g_{2}(\mathcal{X}_1) + \epsilon_3 > 0\}, \quad y_{4} = I\{g_{2}(\mathcal{X}_2) + \epsilon_4 > 0\}.
\end{align*}
We conduct 5-fold cross validation for 10 runs of 10 synthetic datasets of size 10000. 
We compare the predictive performance of CeViT and the baseline ViT model. 
Generation details of synthetic images $(\mathcal{X}_1, \mathcal{X}_2)$ and regression functions are deferred to the Supplement.

\begin{figure}[tb]
    \centering
    \includegraphics[scale = 0.4]{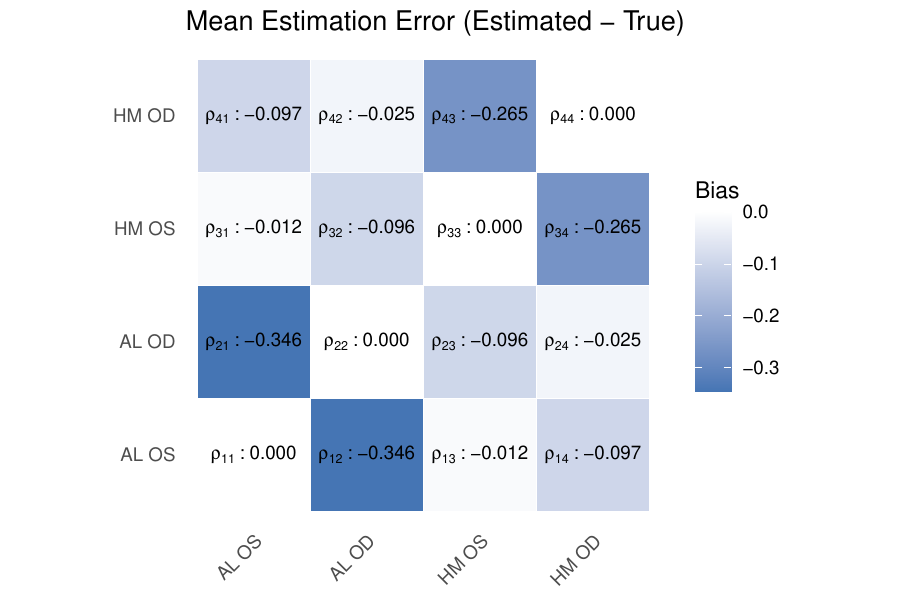}
    \caption{The heatmap of mean estimation error of $\Gamma$ among simulations. }
    \label{fig:simu error}
\end{figure}

\noindent{\textbf{Results.}} 
We first present the results of copula parameter estimation in Figure \ref{fig:simu error}. 
We clearly find that $\rho_{12}$ is strongly underestimated, indicating the occurrence of the stronger covariance phenomenon on regression tasks of $y_1$ and $y_2$. 
We also find that the fMCEM algorithm underestimates $\rho_{34}$ for classification tasks. 
Such an underestimation is a tradeoff between estimation consistency and numerical stability.

Although the copula parameters are underestimated, incorporating the conditional dependence structure significantly improves the predictive performance. 
For evaluation, we compare the predictive performances of models that are trained under the proposed copula loss and the empirical loss respectively. 
For continuous responses (OU AL), we use mean average error (MAE) as the evaluation metric;  
for discrete responses (OU HM), we use the classification accuracy (ACC) and the areas under curve (AUC) as the evaluation metrics.

\begin{figure}[tb]
    \centering
    \includegraphics[scale = .65]{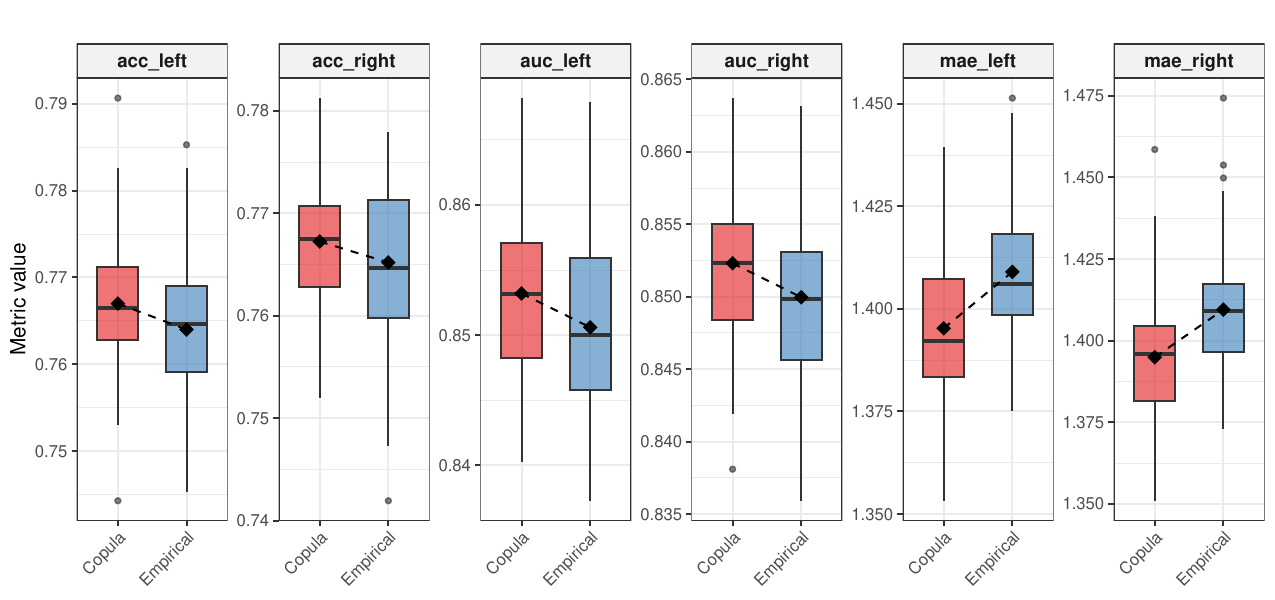}
    \caption{Boxplots of evaluation metrics on synthetic datasets. Copula: models trained under the proposed copula loss; Empirical: models trained under the empirical loss. 
    }
    \label{fig:simu_boxplot}
\end{figure}

The boxplots of the evaluation metrics are presented in Figure \ref{fig:simu_boxplot}. 
We find that CeViT apparently enhances ViT in all metrics. 
We evaluate the significance of improvement through paired $t$-tests, and compute the corresponding Cohen's d to examine whether the improvement is practically meaningful. 
Specifically, we treat those improvement with absolute Cohen's d greater than $0.2$ as substantial improvement. 
As shown by Table \ref{tab: significance simu}, we find the improvement in all metrics are statistically significant and practically substantial. 
These results strongly evidence that the biased estimation of copula parameters still enhance the predictive performances.

\begin{table}[tb]
\centering
\caption{Results on synthetic data: Cohen's $d$; $p$-values in brackets. }
\tiny
\begin{tabular}{lcccccc}
\hline
Comparison & MAE\_L $\downarrow$ & MAE\_R $\downarrow$ & ACC\_L $\uparrow$ & ACC\_R $\uparrow$ & AUC\_L $\uparrow$ & AUC\_R $\uparrow$ \\ 
\hline 
CeViT vs. ViT     &  $\bm{-1.47}$ ($\bm{<.001}$) &  $\bm{-2.08}$ ($\bm{<.001}$) &  $\bm{0.52}$ ($\bm{<.001}$) &  $\bm{0.36}$ ($\bm{0.014}$) &  $\bm{0.88}$ ($\bm{<.001}$) &  $\bm{1.04}$ ($\bm{<.001}$) \\ 
\hline
\multicolumn{7}{l}{\textit{Note:} Values presented as Cohen's $d$ ($p$-value). Bold indicates $\bm{|d| > 0.2}$ or $\bm{p < 0.05}$.} \\
\multicolumn{7}{l}{$\downarrow$: the lower the better; $\uparrow$: the higher the better.} \\
\end{tabular}
\label{tab: significance simu}
\end{table}

\section{Application on the UWF fundus image dataset}
\label{sec: app}
{
In this section, we apply CeViT to our annotated UWF fundus image dataset for OU HM diagnosis and AL predictions. 
Given that ViT outperforms general CNNs in most computer vision tasks, we only present results obtained under ViT-variant models, and omit the results of CNNs.  

\noindent \textbf{Dataset.} The Eye and ENT Hospital of Fudan University has collected 5228 UWF fundus images from the eyes of 2614 patients using the Optomap Daytona scanning laser ophthalmoscope (Daytona, Optos, UK), from December 2014 to June 2020. 
All enrolled patients sought refractive surgery treatment and were exclusively myopia patients. 
The UWF fundus images obtained during the study are exported in 3-channel JPEG format and compressed to a resolution of 224 x 224 pixels to facilitate subsequent analysis.

\noindent{\textbf{Methods for comparison.}} We compare the performances of several models:
\begin{itemize}
    \item \textbf{ViT}: the ViT-base backbone trained under the empirical loss \eqref{empirical loss}. 
    \item \textbf{ViT-A}: the proposed bi-channel ViT with adapters trained under the empirical loss \eqref{empirical loss}.
    \item \textbf{CeViT}: the ViT-base backbone trained under the proposed four-dim copula loss \eqref{copula loss}. 
    \item \textbf{CeViT-A}: the proposed bi-channel ViT with adapters trained under the proposed four-dim copula loss \eqref{copula loss}. 
\end{itemize}

\noindent \textbf{Implementation details} 
For CeViT and CeViT-A, Steps 1 and 3 in Algorithm \ref{alg: CeViT} consist of 25 and 60 epochs to guarantee convergence. 
A batch size of 48 was used throughout the training on all models. 
We utilize the Adam optimizer and employed PyTorch's StepLR as the learning rate scheduler. 
For ViT training, the initial learning rate is set to $10^{-3}$, which decay by a factor of $0.9$ every $4$ epochs. 
In Step 3 of Algorithm \ref{alg: CeViT} for the training of CeViT and CeViT-A,  the initial learning rate is adjusted to $10^{-4}$, decaying by a factor of 0.9 every two epochs.
We conduct parallel training using two RTX 4090 GPUs, and the entire training process takes approximately 4.5 hours to complete.

\noindent{\textbf{Estimation of the correlation matrix}.} 
The results of the estimation of $\Gamma$ given by fMCEM algorithm under the UWF fundus image dataset are presented in Figure \ref{fig:fMCEM_real}. 
The heat map clearly reveals the existence of non-negligible inter-correlation between OU AL and OU HM.
Furthermore, owing to the aforementioned underestimation on regression tasks, we suspect that the conditional correlation between OU AL is pretty high. 
A possible reason is that the OU AL are measured on the same machine at one time, indicating that they may share the same measurement error. 
Meanwhile, we also find that the inter-correlation is much stronger than the intra-correlation between AL and HM for a single eye. 
The boxplot reflects the variation of the estimates of the off-diagonal elements of $\Gamma$ in cross validation, where we find the computation of fMCEM is stable against the random shuffle of training data.

\begin{figure}[tb]
    \centering
\includegraphics[scale = .5]{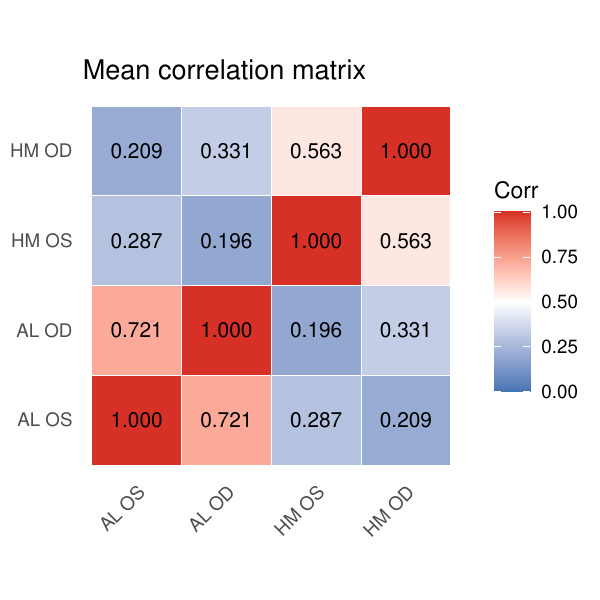}
\includegraphics[scale = .45]{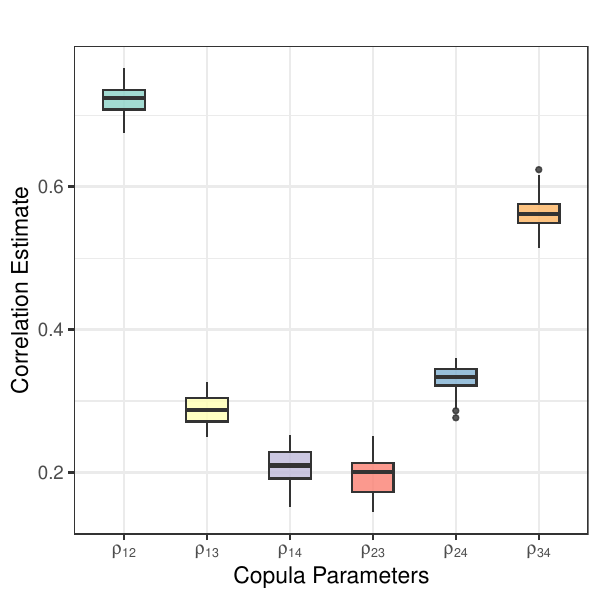}
    \caption{Left: the mean of the estimated correlation matrix $\hat{\Gamma}$ among cross validations; right: boxplots of the estimates of copula parameters among cross validation. }
    \label{fig:fMCEM_real}
\end{figure}

\begin{figure}[tb]
    \centering
    \includegraphics[scale = .4]{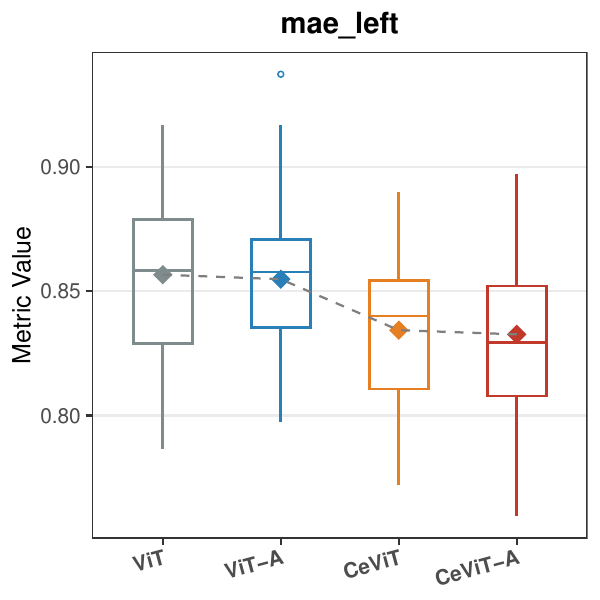}
      \includegraphics[scale = .4]{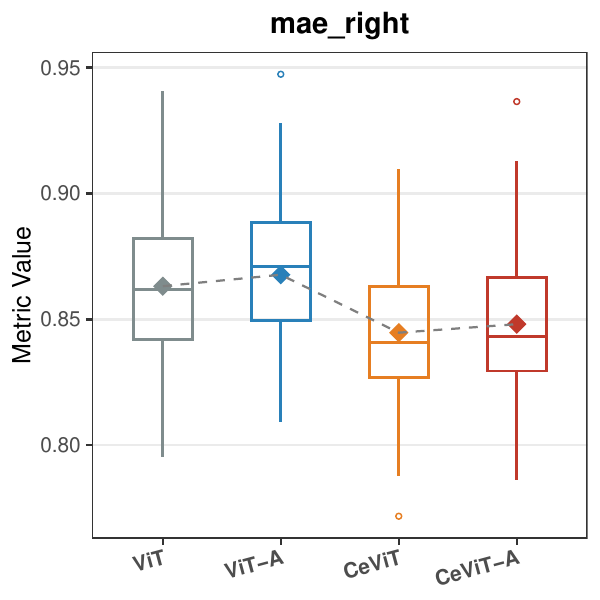}

       \includegraphics[scale = .4]{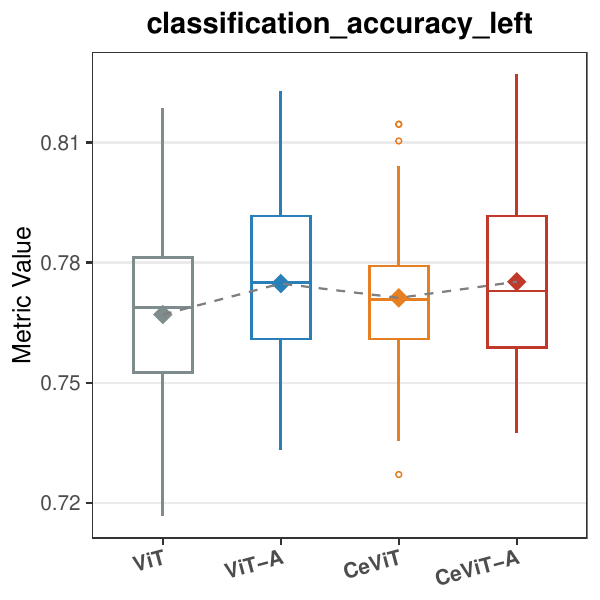}
      \includegraphics[scale = .4]{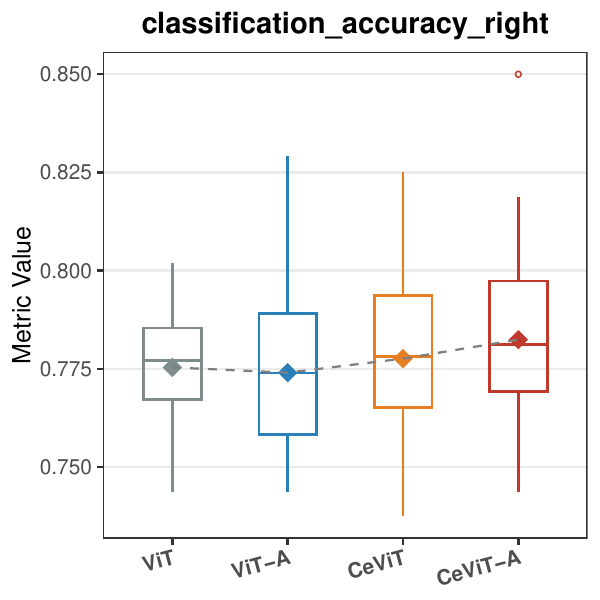}

       \includegraphics[scale = .4]{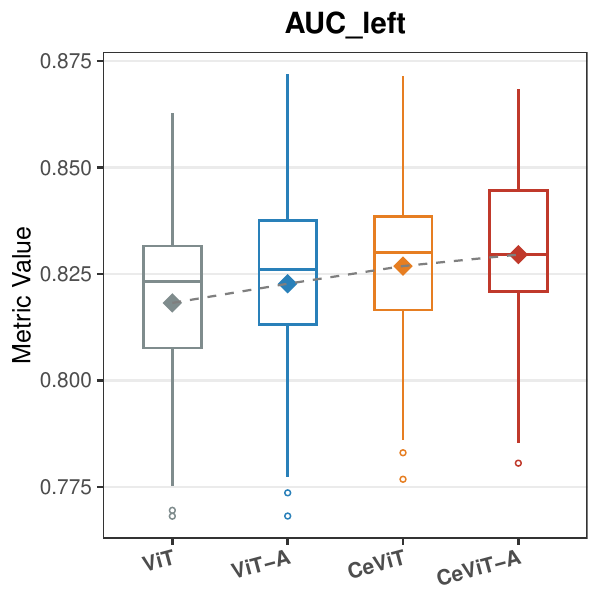}
      \includegraphics[scale = .4]{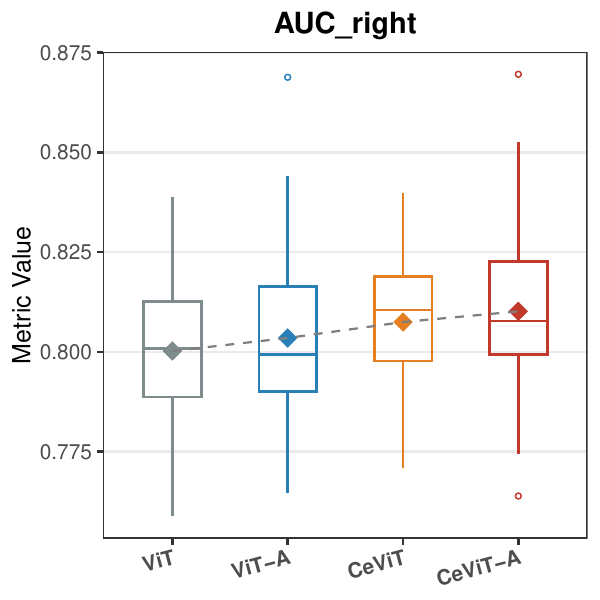}
    \caption{Boxplots of evaluation metrics on the UWF fundus image dataset across 10 runs of 5-fold cross validation. }
    \label{fig: box plot real}
\end{figure}

\noindent \textbf{Result 1: copula loss significantly enhances regression and classification tasks.} 
Figure \ref{fig: box plot real} presents the boxplots of evaluation metrics across 10 runs of 5-fold cross-validation.
We visualize the boxplots of evaluation metrics across the cross-validation experiments in Figure \ref{fig: box plot real}. 
To rigorously quantify the difference between the metrics given by different models, we present pair-model Cohen's d and the $p$-values of paired $t$-tests of ablation experiments in Table \ref{tab: significance REAL}. 
From the boxplots, we find that the proposed copula loss enhances all the regression and classification task under two different backbones. 
The last two lines of Table \ref{tab: significance REAL} demonstrates that the improvement gained by the copula loss is statistically significant and practically meaningful to both regression and classification tasks among all runs of cross validation. 
This result extends the result of CeCNN in that OU HM and AL exhibit non-negligible conditional dependence given the OU UWF fundus images.

\noindent \textbf{Result 2: residual adapters enhance the classification performance.}
From the boxplots, we find the imposing residual adapters enhance both the classification accuracy and AUC. 
The Cohen's d validates that such an enhancement is meaningful. 
Nonetheless, we find that the adapters do NOT benefit regression tasks. 
We conjecture that the inter-ocular asymmetry has substantial impacts on HM, while mildly influences AL. 
For illustration, we present the contour plots of OU t-SNE, categorized by the corresponding HM labels in Figure \ref{fig:oueye_hm_tsne}. 
The contour plot shows the existence of discrepancy between the t-SNE distributions of OU non-HM cases. 
This result indicates that the inter-ocular asymmetry is more likely to occur on non-HM individuals.
Furthermore, we argue that the residual adapters may be underfitting due to the limited training data size. 
As the data size increases in the future, it is anticipated that the residual adapters will become more powerful to enhance the classification performance.

\begin{table}[!htb]
\centering
\caption{Results of pair-wise model comparison: Cohen's $d$; $p$-values in brackets. }
\label{tab:pairwise_results}
\tiny
\begin{tabular}{lcccccc}
\hline
Comparison & MAE\_L $\downarrow$ & MAE\_R $\downarrow$ & ACC\_L $\uparrow$ & ACC\_R $\uparrow$ & AUC\_L $\uparrow$ & AUC\_R $\uparrow$ \\ 
\hline 
CeViT-A vs. CeViT &  -0.09 (0.533) &  $\bm{0.21}$ (0.137) &  $\bm{0.25}$ (0.087) &  $\bm{0.32}$ ($\bm{0.031}$) &  $\bm{0.23}$ (0.111) &  $\bm{0.21}$ (0.143) \\ 
ViT-A vs. ViT     &  -0.06 (0.659) &  $\bm{0.21}$ (0.141) &  $\bm{0.44}$ ($\bm{0.003}$) &  -0.09 (0.535) &  $\bm{0.34}$ ($\bm{0.020}$) &  $\bm{0.23}$ (0.110) \\ 
CeViT vs. ViT     &  $\bm{-1.91}$ ($\bm{<.001}$) &  $\bm{-1.59}$ ($\bm{<.001}$) &  $\bm{0.29}$ ($\bm{0.048}$) &  0.16 (0.269) &  $\bm{1.57}$ ($\bm{<.001}$) &  $\bm{1.33}$ ($\bm{<.001}$) \\ 
CeViT-A vs. ViT-A &  $\bm{-1.40}$ ($\bm{<.001}$) &  $\bm{-1.42}$ ($\bm{<.001}$) &  0.03 (0.847) &  $\bm{0.86}$ ($\bm{<.001}$) &  $\bm{1.09}$ ($\bm{<.001}$) &  $\bm{1.37}$ ($\bm{<.001}$) \\ 
\hline
\multicolumn{7}{l}{\textit{Note:} Values presented as Cohen's $d$ ($p$-value). Bold indicates $\bm{|d| > 0.2}$ or $\bm{p < 0.05}$.} \\
\multicolumn{7}{l}{$\downarrow$: the lower the better; $\uparrow$: the higher the better.} \\
\end{tabular}
\label{tab: significance REAL}
\end{table}

\begin{figure}
    \centering
\includegraphics[width=0.7\linewidth]{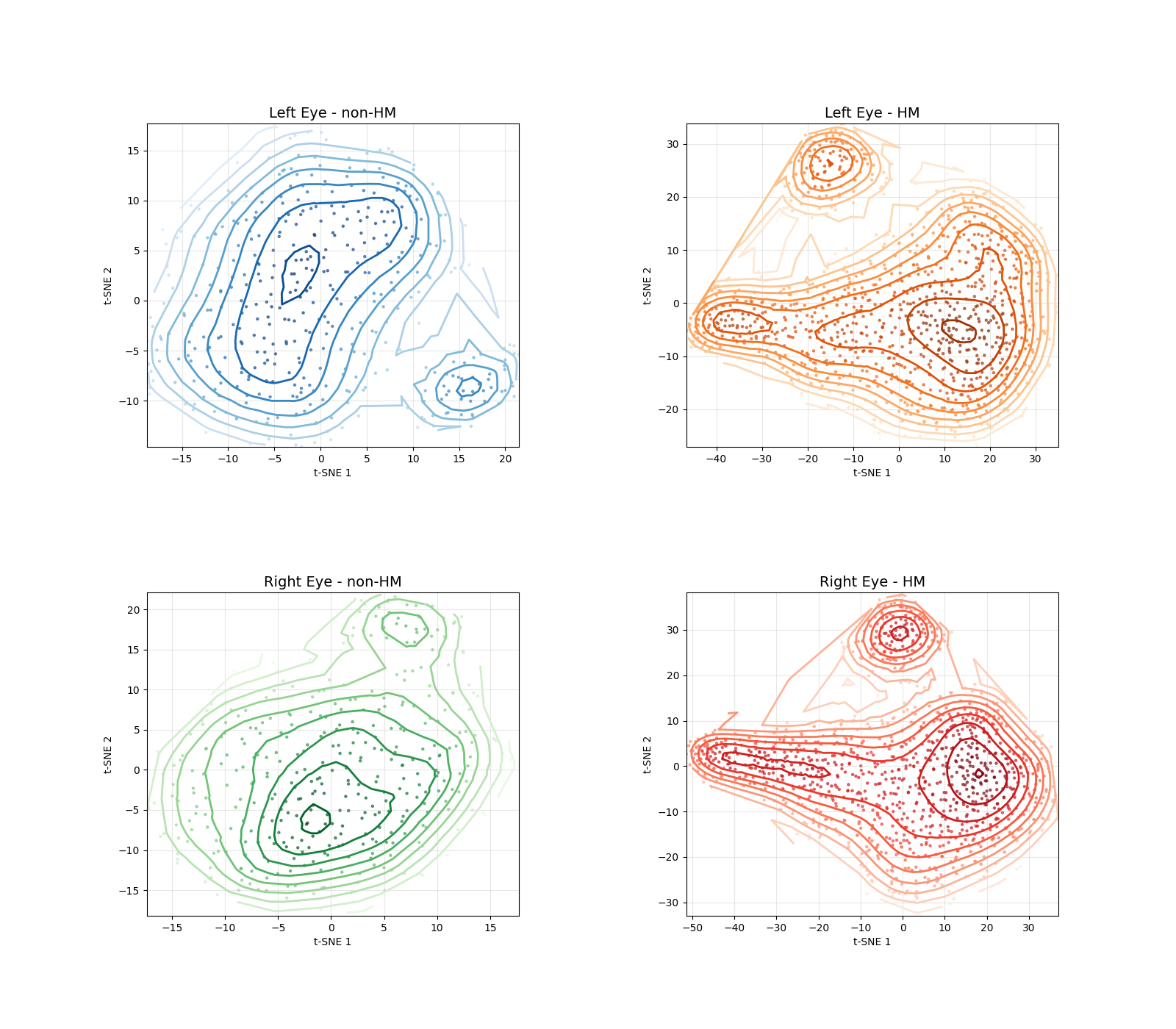}
    \caption{Contour plot of OU t-SNE categorized by HM status. }
    \label{fig:oueye_hm_tsne}
\end{figure}




}

\section{Discussion}
\label{sec: discussion}
In this paper, to jointly predict OU HM and AL based on OU UWF fundus images, we propose a novel ophthalmology AI model CeViT, which is trained under the proposed copula loss. 
The copula parameters in the loss is estimated via a computationally efficient and numerically stable fMCEM algorithm.
The algorithm comprehensively captures the conditional dependence among mixed-type multiple responses, and hence enhances the multitask learning performance of downstream tasks of ViT.

CeViT substantially advances existing AI models for myopia screening.
We illustrate the advancement by comparing with our previous investigations, OU-Copula \citep{li2024oucopula} and CeCNN \citep{zhong2025cecnn}. 
Compared to the pilot OU-Copula, the architecture of CeViT presents a dual-adapter scheme, where the residual adapters introduce a bi-channel architecture to ViT to learn the OU similarity and heterogeneity simultaneously, and the low-rank adapters mitigate overfitting on limited training data.

It is worth noting the advancement from CeCNN to CeViT.
\begin{itemize}
    \item \textit{More general multitask learning loss}.
    Driven by the OU setting, CeViT extends the bivariate copula likelihood loss of CeCNN to higher dimensional cases. 
The presented copula loss in this paper is general to cover $p$-dimensional mixed-type responses for $p \ge 2$, including the bivariate case of CeCNN.

\item \textit{Upgraded copula parameter estimation}.
The increasing dimensionality on mixed-type data calls for new estimation of the copula parameters. 
In the bivariate setting, CeCNN simply estimates the copula parameters through the sample correlation between the Gaussian score of the classification output and the regression residuals.  
However, such a deterministic estimator ignores the variation of latent variables on classification tasks and therefore leads to severely underestimated $\hat{\rho}_{34}$ for classification tasks such that $|\hat{\rho}_{34}| \approx0$. 
In contrast, the proposed fMCEM algorithm extracts the correlation from latent variables directly, and hence only slightly underestimates $\rho_{34}$ but maintains sufficient correlation information to enhance the prediction. 

\item \textit{More comprehensive concerns of overfitting problems}. 
The CeCNN training was once disturbed by the stronger covariance phenomenon mentioned in Section \ref{sec: overfitting}. 
As a result, if CeCNN employs the full ResNet18 backbone, the estimated correlation coefficient between the residuals of the training set on regression tasks will vanish to zero. 
Therefore, CeCNN has to reduce the model size of the backbone to remedy the underestimation on regression tasks, but 
sacrifices the predictive capability. 
In contrast, CeViT conducts fine tuning on a pretrained ViT model and avoids that the underestimated correlation coefficient vanishes to zero on regression tasks.
On classification tasks, the overestimation is naturally balanced by the underestimated fMCEM algorithm, whose numerical stability is guaranteed by  
Proposition \ref{prop: no over 1}.  
\end{itemize}

Despite the inconsistency, the predictive performance of CeViT remains compelling, which is consistent with the existing literature on regression models using the MLE, where inconsistent covariance estimation does not alter the asymptotic predictive performance \citep{zhang2004inconsistent, zhang2023low, li2024fixed}. 
This is theoretically grounded in the fact that the global minimum of the copula loss \eqref{copula loss} and that of the empirical loss \eqref{empirical loss} coincide, since the location of the MLE of a multivariate Gaussian distribution does not depend on the covariance structure. 
In this sense, the estimated copula parameters shape the curvature of the likelihood surface but do not affect the final location of the MLE.

The presented methodology is NOT restricted to OU high myopia diagnosis; 
indeed, it provides a framework applicable to general mixed-type multitask learning scenarios involving multi-domain image covariates that may be correlated. 
While the empirical performance of CeViT is compelling, a comprehensive theoretical understanding remains an open and challenging problem.
The future work may focus on establishing the excess risk bound under the copula loss in both asymptotic and non-asymptotic settings, which deserve separate work.


\bigskip
\begin{center}
{\large\bf SUPPLEMENTARY MATERIAL}

\end{center}

\begin{description}

\item[Title:] Supplementary materials for ``Copula-enhanced Vision Transformer for high myopia diagnosis through OU UWF fundus images" (PDF file).

\end{description}

\bibliographystyle{apalike}
\selectfont
\bibliography{ref}
\end{document}


\def\spacingset#1{\renewcommand{\baselinestretch}%
{#1}\small\normalsize} \spacingset{1}


\if1\blind
{
  \title{\bf Robust prediction of survival outcomes through unified Bayesian analysis of nonparametric transformation models}
  \author{Chong Zhong}
  \maketitle
} \fi

\if0\blind
{
  \bigskip
  \bigskip
  \bigskip
  \begin{center}
    {\LARGE \bf Supplementary materials for ``Copula-enhanced Vision Transformer for high myopia diagnosis through OU UWF fundus images" }
\end{center}
  \medskip
} \fi

\renewcommand*{\thesection}{\Alph{section}}
\renewcommand*{\thesubsection}{\Alph{section}.\arabic{subsection}}
\setcounter{section}{0}

\spacingset{1.9} 
\tableofcontents

\section{Technical proofs}


\subsection{Proof of Theorem 2.1}

\begin{proof}
By the standard factorization of a multivariate Gaussian,
\[
  f(\mathbf{u}) = f(u_1, u_2) \cdot f(u_3, u_4 \mid u_1, u_2),
\]
where $(u_1, u_2)^\top \sim \mathrm{MVN}_2(\mathbf{0}_2, \Gamma_{11})$
and
\[
  (u_3, u_4)^\top \mid (u_1, u_2)^\top
  \sim \mathrm{MVN}_2(\tilde{\bm{\mu}}, \tilde{V}),
\]
with
\[
  \tilde{\bm{\mu}}
  = \Gamma_{21}\Gamma_{11}^{-1}(u_1, u_2)^\top
  =: (\tilde{\mu}_1, \tilde{\mu}_2)^\top,
  \qquad
  \tilde{V} = \Gamma_{22} - \Gamma_{21}\Gamma_{11}^{-1}\Gamma_{12}.
\]
The marginal log-density of $(u_1, u_2)^\top
\sim \mathrm{MVN}_2(\mathbf{0}_2, \Gamma_{11})$ gives the
continuous part
\begin{equation}\label{eq:cont_part}
  \ell_{\mathrm{cont}}
  = -\frac{1}{2(1-\rho_{12}^2)}
    \left[
      \frac{(y_1 - \mu_1)^2}{\sigma_1^2}
      - \frac{2\rho_{12}(y_1-\mu_1)(y_2-\mu_2)}{\sigma_1\sigma_2}
      + \frac{(y_2-\mu_2)^2}{\sigma_2^2}
    \right] + C_1.
\end{equation}

From the latent variable representation~(8), for $k \in \{3, 4\}$,
define the threshold $c_k := \Phi^{-1} \circ \mathcal{S}(-\mu_k)$,
so that
\[
  \Pr(y_k = 1 \mid u_1, u_2) = \Pr(u_k \geq c_k \mid u_1, u_2),
  \qquad
  \Pr(y_k = 0 \mid u_1, u_2) = \Pr(u_k < c_k \mid u_1, u_2).
\]
Since $(u_3, u_4)^\top \mid (u_1,u_2)^\top \sim
\mathrm{MVN}_2(\tilde{\bm{\mu}}, \tilde{V})$, we derive
$\Pr(y_3, y_4 \mid u_1, u_2)$ case by case.

\textbf{Case 1: $y_3 = 0,\; y_4 = 0$.}
\[
  \Pr(u_3 < c_3,\, u_4 < c_4 \mid u_1, u_2)
  = \Phi_2\!\left\{(c_3, c_4)^\top;\, \tilde{\bm{\mu}},\, \tilde{V}\right\}
  = \Phi_2\!\left\{(-\mu_3,-\mu_4)^\top;\, \tilde{\bm{\mu}},\, \tilde{V}\right\}.
\]

\textbf{Case 2: $y_3 = 1,\; y_4 = 1$.}
\begin{align*}
  \Pr(u_3 \geq c_3,\, u_4 \geq c_4 \mid u_1, u_2)
  &= 1 - \Pr(u_3 < c_3 \mid u_1,u_2)
       - \Pr(u_4 < c_4 \mid u_1,u_2)\\
       &+ \Pr(u_3 < c_3,\, u_4 < c_4 \mid u_1,u_2) 
  = \Phi_2\!\left\{(-c_3,-c_4)^\top;\, \tilde{\bm{\mu}},\, \tilde{V}\right\}.
\end{align*}

\textbf{Case 3: $y_3 = 1,\; y_4 = 0$.}
\begin{align*}
  \Pr(u_3 \geq c_3,\, u_4 < c_4 \mid u_1, u_2)
  &= \Pr(u_4 < c_4 \mid u_1,u_2)
   - \Pr(u_3 < c_3,\, u_4 < c_4 \mid u_1,u_2) \\
  &= \Phi\!\left\{\frac{c_4 - \tilde{\mu}_2}{\sqrt{\tilde{V}_{22}}}\right\}
   - \Phi_2\!\left\{(c_3, c_4)^\top;\, \tilde{\bm{\mu}},\, \tilde{V}\right\} \\
  &= \Phi_{\tilde{\mu}_2,\tilde{V}_{22}}\!\left(\Phi^{-1}\circ\mathcal{S}(-\mu_4)\right)
   - \Phi_2\!\left\{(-\mu_3,-\mu_4)^\top;\, -\tilde{\bm{\mu}},\, \tilde{V}\right\}.
\end{align*}

\textbf{Case 4: $y_3 = 0,\; y_4 = 1$.}
\begin{align*}
  \Pr(u_3 < c_3,\, u_4 \geq c_4 \mid u_1, u_2)
  &= \Pr(u_3 < c_3 \mid u_1,u_2)
   - \Pr(u_3 < c_3,\, u_4 < c_4 \mid u_1,u_2) \\
  &= \Phi\!\left\{\frac{c_3 - \tilde{\mu}_1}{\sqrt{\tilde{V}_{11}}}\right\}
   - \Phi_2\!\left\{(c_3, c_4)^\top;\, \tilde{\bm{\mu}},\, \tilde{V}\right\} \\
  &= \Phi_{\tilde{\mu}_1,\tilde{V}_{11}}\!\left(\Phi^{-1}\circ\mathcal{S}(-\mu_3)\right)
   - \Phi_2\!\left\{(-\mu_3,-\mu_4)^\top;\, -\tilde{\bm{\mu}},\, \tilde{V}\right\}.
\end{align*}

Combining all four cases, the binary conditional log-probability is
\begin{align*}
  \ell_{\mathrm{bin}|\mathrm{cont}}
  = \log\Big[
    &\; y_4(1-2y_3)\,
       \Phi_{\tilde{\mu}_1,\tilde{V}_{11}}\!\left(\Phi^{-1}\circ\mathcal{S}(-\mu_3)\right)
    + y_3(1-2y_4)\,
       \Phi_{\tilde{\mu}_2,\tilde{V}_{22}}\!\left(\Phi^{-1}\circ\mathcal{S}(-\mu_4)\right) \\
    &+ (1-2y_3)(1-2y_4)\,
       \Phi_2\!\left\{(-\mu_3,-\mu_4)^\top;\,\tilde{\bm{\mu}},\,\tilde{V}\right\}
    + y_3 y_4
  \Big].
\end{align*}

Adding $\ell_{\mathrm{cont}}$ from~\eqref{eq:cont_part} and the
normalization constant $C$ yields equation~(10), completing the
proof.
\end{proof}

The following corollary extends Theorem 2.1 to general $p$-dimensional Gaussian copula by using the inclusion-exclusion principle.

\begin{corollary}[General $p$-dimensional copula loss]
\label{cor:general}
Let $\mathbf{y} = (y_1,\ldots,y_p)^\top$ with $p_0$ continuous
responses $y_1,\ldots,y_{p_0} \in \mathbb{R}$ and $p_1 = p - p_0$
binary responses $y_{p_0+1},\ldots,y_p \in \{0,1\}$.
Let $\mathcal{A} = \{k \in \{p_0{+}1,\ldots,p\} : y_k = 1\}$
be the index set of positive binary responses. 
Let $T \subseteq \mathcal{A}$ be the flipped index set such that 
$\bm{b}^{(T)} = (b_{p_0+1}^{T}, \ldots, b_p^{(T)})$, where 
\begin{equation}\label{eq:limits}
  b_k^{(T)} =
  \begin{cases}
    \Phi^{-1}{\circ}\,\mathcal{S}(-\mu_k)
      & k \notin \mathcal{A}
        \quad (y_k = 0),\\
    -\Phi^{-1}{\circ}\,\mathcal{S}(-\mu_k)
    = \Phi^{-1}{\circ}\,\mathcal{S}(\mu_k)
      & k \in \mathcal{A}\setminus T
        \quad (y_k = 1,\text{ not flipped}),\\
    \Phi^{-1}{\circ}\,\mathcal{S}(-\mu_k)
      & k \in T
        \quad (\text{flipped by inclusion--exclusion}).
  \end{cases}
\end{equation}
Under the Gaussian copula
$\mathbf{u} = (e_1,\ldots,e_{p_0}, u_{p_0+1},\ldots,u_p)^\top
\sim \mathrm{MVN}_p(\mathbf{0}_p, \Gamma)$,
the joint log-density of
$\mathbf{y}|\mathbf{X}$ is
\[
  \ell(\mathbf{y}|\mathbf{X})
  = \ell_{\mathrm{cont}} + \ell_{\mathrm{bin}|\mathrm{cont}} + C,
\]
where
\begin{equation}\label{eq:cont_gen}
  \ell_{\mathrm{cont}}
  = -\tfrac{1}{2}\mathbf{e}^\top \Gamma_{11}^{-1}\mathbf{e}
    -\tfrac{1}{2}\log|\Gamma_{11}| + C_1,
  \qquad
  \mathbf{e} = (e_1,\ldots,e_{p_0})^\top,
\end{equation}
and
\begin{equation}\label{eq:bin_gen}
  \ell_{\mathrm{bin}|\mathrm{cont}}
  = \log\!\left[
      \sum_{T \subseteq \mathcal{A}}
      (-1)^{|T|}\,
      \Phi_{p_1}\!\left(
        \mathbf{b}^{(T)};\;
        \boldsymbol{\mu}_e,\,
        V_e
      \right)
    \right],
\end{equation}
with conditional mean and variance
\[
  \boldsymbol{\mu}_e
  = \Gamma_{21}\Gamma_{11}^{-1}\mathbf{e},
  \qquad
  V_e = \Gamma_{22} - \Gamma_{21}\Gamma_{11}^{-1}\Gamma_{12}.
\]

\end{corollary}

\begin{proof}
The continuous part~\eqref{eq:cont_gen} follows directly from
the marginal density of $(u_1,\ldots,u_{p_0})^\top
\sim \mathrm{MVN}_{p_0}(\mathbf{0}, \Gamma_{11})$.

Given $(u_1,\ldots,u_{p_0})$, the conditional distribution of
the latent binary scores is
\[
  (u_{p_0+1},\ldots,u_p)^\top \mid (u_1,\ldots,u_{p_0})
  \sim \mathrm{MVN}_{p_1}(\boldsymbol{\mu}_e, V_e).
\]
From the latent variable representation~(8), for each
$k \in \{p_0+1,\ldots,p\}$, define the threshold
$c_k := \Phi^{-1}{\circ}\,\mathcal{S}(-\mu_k)$, so that
\begin{align*}
    \Pr(y_k = 1 \mid u_1,\ldots,u_{p_0})
  = \Pr(u_k \geq c_k \mid u_1,\ldots,u_{p_0}),\\
  \Pr(y_k = 0 \mid u_1,\ldots,u_{p_0})
  = \Pr(u_k < c_k \mid u_1,\ldots,u_{p_0}).
\end{align*}
Therefore, the conditional probability of the observed binary
vector is
\begin{equation}\label{eq:gen_prob}
  \Pr(y_{p_0+1},\ldots,y_p \mid u_1,\ldots,u_{p_0})
  = \Pr\!\left(
      \bigcap_{k=p_0+1}^{p}
      \{u_k \in R_k\}
      \;\Big|\; u_1,\ldots,u_{p_0}
    \right),
\end{equation}
where
\[
  R_k =
  \begin{cases}
    [c_k,\, +\infty) & y_k = 1,\\
    (-\infty,\, c_k) & y_k = 0.
  \end{cases}
\]
For each $k \notin \mathcal{A}$ (i.e.\ $y_k = 0$), the event
$\{u_k < c_k\}$ is already expressed as an upper-tail CDF region
and requires no further manipulation. For each $k \in \mathcal{A}$
(i.e.\ $y_k = 1$), we write
\[
  \Pr(u_k \geq c_k) = 1 - \Pr(u_k < c_k),
\]
and apply the inclusion--exclusion principle over all subsets
$T \subseteq \mathcal{A}$ to expand the intersection in
\eqref{eq:gen_prob}:
\begin{align}\label{eq:incl_excl}
  &\Pr\!\left(
    \bigcap_{k \notin \mathcal{A}} \{u_k < c_k\}
    \;\cap\;
    \bigcap_{k \in \mathcal{A}} \{u_k \geq c_k\}
    \;\Big|\; u_1,\ldots,u_{p_0}
  \right) \notag\\
  &= \sum_{T \subseteq \mathcal{A}}
     (-1)^{|T|}
     \Pr\!\left(
       \bigcap_{k \notin \mathcal{A}} \{u_k < c_k\}
       \;\cap\;
       \bigcap_{k \in \mathcal{A}\setminus T} \{u_k \geq c_k\}
       \;\cap\;
       \bigcap_{k \in T} \{u_k < c_k\}
       \;\Big|\; u_1,\ldots,u_{p_0}
     \right).
\end{align}
Each term inside the sum is now a probability of the form
$\Pr(\mathbf{u}_{(p_1)} \leq \mathbf{b}^{(T)})$ under
$\mathrm{MVN}_{p_1}(\boldsymbol{\mu}_e, V_e)$,
where the coordinate-wise upper limits $b_k^{(T)}$ are given
by~\eqref{eq:limits}. Specifically:
\begin{itemize}
  \item For $k \not \in \mathcal{A}$ ($y_k = 0$): the region is $\{u_k < c_k\}$, so $b_k^{(T)} = c_k =
        \Phi^{-1}{\circ}\,\mathcal{S}(-\mu_k)$.
  \item For $k \in \mathcal{A}\setminus T$ ($y_k = 1$, not
        flipped): the region is $\{u_k \geq c_k\}$, which under
        the symmetric Gaussian corresponds to an upper limit of
        $-c_k = \Phi^{-1}{\circ}\,\mathcal{S}(\mu_k)$.
  \item For $k \in T$ ($y_k = 1$, flipped by the
        inclusion--exclusion): the region becomes $\{u_k < c_k\}$,
        so $b_k^{(T)} = c_k = \Phi^{-1}{\circ}\,\mathcal{S}(-\mu_k)$.
\end{itemize}
Therefore each term equals
$\Phi_{p_1}(\mathbf{b}^{(T)};\boldsymbol{\mu}_e, V_e)$,
which is the CDF of $\mathrm{MVN}_{p_1}(\boldsymbol{\mu}_e, V_e)$
evaluated at $\mathbf{b}^{(T)}$, giving~\eqref{eq:bin_gen}.

\end{proof}

\subsection{Proof of Proposition 1}
\subsubsection{Auxiliary Lemmas}

Let $\Phi(\cdot)$ and $\phi(\cdot)$ denote the standard normal CDF and PDF, 
respectively. Let $\Phi_2(\cdot, \cdot; \rho)$ denote the bivariate standard 
normal CDF with correlation $\rho$.

\begin{lemma}[Slepian's inequality for orthant probabilities]
\label{lem:slepian}
Let $(Z_1, Z_2)^\top \sim N(\mathbf{0}, \Sigma)$ with $\Sigma = \begin{pmatrix} 
1 & \rho \\ \rho & 1 \end{pmatrix}$ and $\rho > 0$. For any thresholds $c_1, c_2$,
\[
Pr\{Z_1 > c_1, Z_2 > c_2\} = \Phi_2(-c_1, -c_2; \rho)
\]
is strictly increasing in $\rho$.
\end{lemma}

The result of Lemma \ref{lem:slepian} is trivial. 
The same result also holds for the lower orthant $Pr\{Z_1 \leq c_1, Z_2 \leq c_2\}$. 

\begin{lemma}[Conditional covariance in truncated bivariate normal]
\label{lem:tallis}
Let $(Z_1, Z_2)^\top \sim N(\mathbf{0}, \boldsymbol{\Sigma})$ with $\boldsymbol{\Sigma} = \begin{pmatrix} 
1 & \rho \\ \rho & 1 \end{pmatrix}$ and $\rho > 0$. Consider the truncated 
region $\mathcal{R} = (c_1, \infty) \times (c_2, \infty)$. Let 
\[
\alpha(c_1, c_2) = P(Z_1 > c_1, Z_2 > c_2) = \Phi_2(-c_1, -c_2; \rho)
\]
be the joint survival function. Then:
\begin{enumerate}
\item The conditional covariance admits the representation
\[
\mathrm{Cov}(Z_1, Z_2 \mid \mathcal{R}) 
= \rho - \frac{\partial^2\log \alpha(c_1 , c_2 )}{\partial t_1 \partial t_2} 
. 
\]
\item For $\rho > 0$ and finite $c_1, c_2$, we have 
$\mathrm{Corr}(Z_1, Z_2 \mid \mathcal{R}) > \rho$.
\end{enumerate}
\end{lemma}

\begin{proof}
 The moment generating function (MGF) of 
the truncated distribution with truncation $W_1 > a_1, W_2 > a_2$ is given by \citet[Eq. (2)]{tallis1961moment}:
\[
\alpha m(t_1, t_2) = e^T \Phi_2(b_1, b_2; \rho),
\]
where $T = \frac{1}{2}(t_1^2 + t_2^2 + 2\rho t_1 t_2)$, $b_j = a_j - \zeta_j$, 
and $\zeta_1 = t_1 + \rho t_2$, $\zeta_2 = \rho t_1 + t_2$. Setting 
$a_j = c_j$ and noting that $\alpha = \Phi_2(-c_1, -c_2; \rho)$, we have:
\[
\alpha m(t_1, t_2) = e^{\frac{1}{2}(t_1^2 + t_2^2 + 2\rho t_1 t_2)} 
\Phi_2\big(c_1 - (t_1 + \rho t_2),\; c_2 - (\rho t_1 + t_2);\; \rho\big).
\]
By symmetry of the bivariate normal, $\Phi_2(c_1 - \zeta_1, c_2 - \zeta_2; \rho) 
= \Phi_2(-c_1 + \zeta_1, -c_2 + \zeta_2; \rho) = \alpha(c_1 - \zeta_1, c_2 - \zeta_2)$, 
where $\alpha(\cdot, \cdot)$ is the survival function defined above. 
Taking cross-derivative 
$\partial^2 \log(\alpha m) / \partial t_1 \partial t_2$ at the point $(t_1, t_2) = (0, 0)$, we obtain the covariance conditional on the truncation region $\mathcal{R}$. 
The term from $\partial^2 T / \partial t_1 \partial t_2$ contributes $\rho$, and the term from the survival function contributes the negative derivative of 
$\log \alpha$, giving the stated representation in the first assertion. 

Note that $\frac{\partial^2\log \alpha(c_1 , c_2 )}{\partial t_1 \partial t_2} <0$. 
Therefore, the second assertion holds. 

\end{proof}

\begin{remark}
\label{rmk: inverse stratement}
For the discordant quadrant $\mathcal{R}_{10} = (c_1, \infty) \times 
(-\infty, c_2]$, an analogous representation holds with 
$\mathrm{Corr}(Z_1, Z_2 \mid \mathcal{R}_{10}) < \rho$, and the conditional 
correlation is decreasing in $\mathrm{Cov}(c_1, c_2)$.
\end{remark}



\begin{lemma}[Monotonicity of concordance probability]
\label{lem:conc_mono}
Let $(u_3, u_4)^\top \sim N(\mathbf{0}, \Gamma_{22})$ with $\mathrm{Corr}(u_3, u_4) 
= \rho_{34} > 0$. For thresholds $c_j = -\hat{q}_j$, let $P_{\mathrm{conc}}(\hat{q}_3, \hat{q}_4) = P(y_3 = y_4 \mid \hat{q}_3, \hat{q}_4)$ be the probability of 
labels $(y_3, y_4)$ are concordant. 
We have  $\mathbb{E}(P_{\mathrm{conc}})$ is strictly increasing regarding to
$\gamma := \mathrm{Cov}(\hat{q}_3, \hat{q}_4)$, given that the marginal distributions of $\hat{q}_3, \hat{q}_4$ fixed.
\end{lemma}

\begin{proof}
We have
\begin{align*}
P_{\mathrm{conc}}(\hat{q}_3, \hat{q}_4) 
&= P(y_3 = 1, y_4 = 1) + P(y_3 = 0, y_4 = 0) \\
&= \Phi_2(\hat{q}_3, \hat{q}_4; \rho_{34}) + \Phi_2(-\hat{q}_3, -\hat{q}_4; \rho_{34}) \\
&= 2\Phi_2(\hat{q}_3, \hat{q}_4; \rho_{34}),
\end{align*}
where the last equality follows from the symmetry $\Phi_2(a, b; \rho) = \Phi_2(-a, -b; \rho)$.

Taking the expectation over the bivariate distribution of $(\hat{q}_3, \hat{q}_4)$ 
with covariance $\gamma$ (and fixed marginal variances), Lemma~\ref{lem:slepian} 
implies that $\mathbb{E}[\Phi_2(\hat{q}_3, \hat{q}_4; \rho_{34})]$ is strictly 
increasing in $\gamma$ for $\rho_{34} > 0$. Hence 
$\frac{\partial}{\partial \gamma} \mathbb{E}[P_{\mathrm{conc}}] > 0$.
\end{proof}

\subsubsection{Proof}
Let $\rho_{34}^* = \mathbb{E}\left[ \mathrm{Corr}(u_3, u_4 \mid \mathcal{R}) \right]$ be the probability limit of $\tilde{\rho}_{34}$, where $\mathcal{R} = \mathcal{R}_{y_3 y_4}$ denotes the truncation region determined by $(y_3, y_4, c_3, c_4)$, and the expectation is over the 
distribution of $(y_3, y_4, c_3, c_4)$.
We have the following decomposition
\[
\rho_{34}^* = w_{11} \rho_{11} + w_{00} \rho_{00} + w_{10} \rho_{10} + w_{01} \rho_{01},
\]
where $w_{ab} = \mathbb{E}[P(y_3 = a, y_4 = b \mid c_3, c_4)]$ and 
$\rho_{ab} = \mathbb{E}[\mathrm{Corr}(u_3, u_4 \mid y_3 = a, y_4 = b, c_3, c_4)]$.
By symmetry of the bivariate normal, $\rho_{11} = \rho_{00}$ and $\rho_{10} = \rho_{01}$.
Similarly, $w_{11} = w_{00}$ and $w_{10} = w_{01}$ when the thresholds have 
symmetric marginal distributions (which holds asymptotically for balanced data).
The following proposition reveals the concordance dominance under the stronger covariance phenomenon.

\begin{proposition}[Concordance dominance]
\label{prop:conc_dom}
Assume the baseline concordance condition:
\[
\mathbb{E}[P(y_3 = y_4 \mid \mathcal{X}_1, \mathcal{X}_2)] > \frac{1}{2},
\]
i.e., concordant pairs are more frequent than discordant pairs under the true 
data generating process. If the higher covariance phenomenon occurs on $y_3$ and 
$y_4$, then under the estimated thresholds,
\[
w_{11} + w_{00} > w_{10} + w_{01}.
\]
\end{proposition}

\begin{proof}
The concordance dominance condition is $w_{11} + w_{00} > 1/2$. Under the true 
parameters, $\gamma_0 = \mathrm{Cov}(q_3, q_4)$, we have 
$\mathbb{E}[P_{\mathrm{conc}}(q_3, q_4)] > 1/2$ by the baseline assumption.

Under the higher covariance phenomenon, $\hat{\gamma} = \mathrm{Cov}(\hat{q}_3, \hat{q}_4) 
> \gamma_0$. By Lemma~\ref{lem:conc_mono},
\[
w_{11} + w_{00} = \mathbb{E}[P_{\mathrm{conc}}(\hat{q}_3, \hat{q}_4)] 
> \mathbb{E}[P_{\mathrm{conc}}(q_3, q_4)] > \frac{1}{2}.
\]
Hence $w_{11} + w_{00} > 1 - (w_{11} + w_{00}) = w_{10} + w_{01}$.
\end{proof}
We are now in the position to prove Proposition 1. 
Note that we only need to show the case where $\rho_{34}>0$ and the case where $\rho_{34} <0$ follows the same proof. 

\begin{proof}
For the \textbf{concordant quadrants} $(1,1)$ and $(0,0)$, Lemma~\ref{lem:tallis} 
establishes:
\begin{itemize}
\item $\rho_{11} = \rho_{00} > \rho_{34}$.
\item $\rho_{11}$ and $\rho_{00}$ are strictly increasing in 
$\mathrm{Cov}(c_3, c_4)$.
\end{itemize}

For the \textbf{discordant quadrants} $(1,0)$ and $(0,1)$, applying 
Lemma~\ref{lem:tallis} to the transformed variables $(u_3, -u_4)$ (which have 
correlation $-\rho_{34}$) and $(-u_3, u_4)$, respectively, yields:
\begin{itemize}
\item $\rho_{10} = \rho_{01} < \rho_{34}$.
\item $\rho_{10}$ and $\rho_{01}$ are decreasing in $\mathrm{Cov}(c_3, c_4)$.
\end{itemize}
Let $\gamma = \mathrm{Cov}(c_3, c_4) = \mathrm{Cov}(\hat{q}_3, \hat{q}_4)$. 
The derivative of $\rho_{34}^*$ with respect to $\gamma$ is
\begin{equation}
\frac{d \rho_{34}^*}{d \gamma} 
= \underbrace{\sum_{ab} w_{ab} \cdot \frac{\partial \rho_{ab}}{\partial \gamma}}
_{\text{within-quadrant}} 
+ \underbrace{\sum_{ab} \frac{\partial w_{ab}}{\partial \gamma} \cdot \rho_{ab}}
_{\text{weight shift}}.
\label{eq:deriv}
\end{equation}

\textbf{Within-quadrant term.} Note that
\[
\frac{\partial \rho_{11}}{\partial \gamma} = \frac{\partial \rho_{00}}{\partial \gamma} > 0,
\quad
\frac{\partial \rho_{10}}{\partial \gamma} = \frac{\partial \rho_{01}}{\partial \gamma} < 0.
\]
By Proposition \ref{prop:conc_dom} and 
the fact that $\rho_{11} > \rho_{10}$, the positive contributions 
from concordant quadrants outweigh the negative ones from discordant quadrants 
in magnitude. Formally,
\[
\sum_{ab} w_{ab} \cdot \frac{\partial \rho_{ab}}{\partial \gamma}
= 2 w_{11} \cdot \frac{\partial \rho_{11}}{\partial \gamma} 
+ 2 w_{10} \cdot \frac{\partial \rho_{10}}{\partial \gamma} > 0,
\]
because $w_{11} > w_{10}$ and 
$|\partial \rho_{11} / \partial \gamma| > |\partial \rho_{10} / \partial \gamma|$ 
(the concordant conditional correlation is more sensitive to threshold correlation 
than the discordant one, as can be verified from the Tallis derivative formulas).

\textbf{Weight shift term.} By Lemma~\ref{lem:slepian}, for $\rho_{34} > 0$:
\[
\frac{\partial w_{11}}{\partial \gamma} = \frac{\partial w_{00}}{\partial \gamma} > 0,
\quad
\frac{\partial w_{10}}{\partial \gamma} = \frac{\partial w_{01}}{\partial \gamma} < 0.
\]

This shifts probability mass from the low-correlation discordant quadrants 
to the high-correlation concordant quadrants. Since $\rho_{11} > \rho_{10}$,
\[
\sum_{ab} \frac{\partial w_{ab}}{\partial \gamma} \cdot \rho_{ab}
= 2 \frac{\partial w_{11}}{\partial \gamma} (\rho_{11} - \rho_{10}) > 0.
\]

Both terms in \eqref{eq:deriv} are positive, establishing that 
$d \rho_{34}^* / d \gamma > 0$.

Under the stronger covariance phenomenon, the neural network outputs satisfy
$\hat{\gamma} = \mathrm{Cov}(\hat{q}_3, \hat{q}_4) > \gamma_0$. 
Therefore
\[
\mathbb{E}(\tilde{\rho}_{34}) \to \rho_{34}^*(\hat{\gamma}) 
> \rho_{34}^*(\gamma_0) = \rho_{34}.
\]
As $\hat{\gamma} / \gamma_0 \to \infty$, we have $w_{10}, w_{01} \to 0$ 
(the probability of discordant pairs vanishes) and $\rho_{11}, \rho_{00} \to 1$ 
(the concordant conditional correlations approach perfect correlation), 
so $\rho_{34}^* \to 1$.
\end{proof}

\subsection{Proof of Proposition 2}
\begin{proof}
The fMCEM algorithm replaces the latent variables $u_3, u_4$ with their 
conditional expectations given the truncation pattern. For each observation,
\[
\tilde{u}_j = \mathbb{E}[u_j \mid y_j, \hat{q}_j], \quad j = 3,4.
\]

These are the inverse Mills ratios:
\begin{align*}
\tilde{u}_j \mid (y_j = 1) &= \frac{\phi(\hat{q}_j)}{\Phi(\hat{q}_j)} := \lambda_1(\hat{q}_j), 
\tilde{u}_j \mid (y_j = 0) &= -\frac{\phi(\hat{q}_j)}{1 - \Phi(\hat{q}_j)} := \lambda_0(\hat{q}_j).
\end{align*}

\medskip
\noindent\textbf{Step 1: Variance deflation.}
By the law of total variance,
\[
\mathrm{Var}(\tilde{u}_j) = \mathrm{Var}(u_j) - \mathbb{E}[\mathrm{Var}(u_j \mid y_j, \hat{q}_j)]
= 1 - \mathbb{E}[\mathrm{Var}(u_j \mid y_j, \hat{q}_j)].
\]

The conditional variance $\mathrm{Var}(u_j \mid y_j, \hat{q}_j)$ is the variance 
of a truncated standard normal. For any finite truncation threshold, the 
truncated variance satisfies $\mathrm{Var}(u_j \mid y_j, \hat{q}_j) \in (0, 1]$, 
with equality only for no truncation ($\hat{q}_j \to \infty$ for $y_j = 1$, or 
$\hat{q}_j \to -\infty$ for $y_j = 0$). In practice, the thresholds $\hat{q}_j$ 
are finite with probability one, so
\[
\mathrm{Var}(\tilde{u}_j) \leq 1 - v_{\min} =: \tau^2 < 1,
\]
where $v_{\min} = \inf_{\hat{q}} \mathrm{Var}(u_j \mid y_j, \hat{q}) > 0$.

\medskip
\noindent\textbf{Step 2: Heterogeneity across quadrants.}
The fMCEM estimator computes the sample correlation of 
$\{(\tilde{u}_{i3}, \tilde{u}_{i4})\}_{i=1}^n$. Even as 
$\mathrm{Cov}(\hat{q}_3, \hat{q}_4) \to \infty$, the imputed values from 
different $(y_3, y_4)$ quadrants exhibit fundamentally different behavior:

\begin{itemize}
\item For concordant quadrants $(1,1)$ and $(0,0)$: $\tilde{u}_{i3}$ and 
$\tilde{u}_{i4}$ have the same sign and are perfectly correlated as 
$\mathrm{Cov}(\hat{q}_3, \hat{q}_4) \to \infty$.
\item For discordant quadrants $(1,0)$ and $(0,1)$: $\tilde{u}_{i3}$ and 
$\tilde{u}_{i4}$ have opposite signs.
\end{itemize}

In the limit $\mathrm{Cov}(\hat{q}_3, \hat{q}_4) \to \infty$, the discordant 
quadrant probabilities $w_{10}$ and $w_{01}$ approach zero but do not vanish 
entirely for finite $n$ unless $\rho_{34} = 1$. The presence of any discordant 
observations creates an irreducible attenuation in the overall sample 
correlation.

\medskip
\noindent\textbf{Step 3: Characterization of the upper bound.}
The fMCEM estimator converges to
\[
\rho^{\mathrm{fMCEM}} = 
\frac{\sum_{a,b} w_{ab} \cdot \mathbb{E}[\tilde{u}_3 \tilde{u}_4 \mid y_3=a, y_4=b]}
{\sqrt{\sum_a w_{a\cdot} \mathbb{E}[\tilde{u}_3^2 \mid y_3=a] 
\cdot \sum_b w_{\cdot b} \mathbb{E}[\tilde{u}_4^2 \mid y_4=b]}}.
\]

Let $\gamma = \mathrm{Cov}(\hat{q}_3, \hat{q}_4) \to \infty$. In this limit,
\begin{itemize}
\item $\mathbb{E}[\tilde{u}_3^2 \mid y_3] \to \mathbb{E}[\tilde{u}_3^2 \mid y_3] < 1$ 
(variance deflation persists).
\item $\mathbb{E}[\tilde{u}_3 \tilde{u}_4 \mid y_3 = y_4] \to 
\mathbb{E}[\tilde{u}_3^2 \mid y_3]$ (perfect within-quadrant correlation).
\item $\mathbb{E}[\tilde{u}_3 \tilde{u}_4 \mid y_3 \neq y_4] \to 
-\mathbb{E}[\tilde{u}_3^2 \mid y_3]$ (perfect within-quadrant negative correlation).
\end{itemize}

Thus,
\[
\rho^{\mathrm{fMCEM}} \to 
\frac{(w_{11} + w_{00}) \tau^2 - (w_{10} + w_{01}) \tau^2}
{(w_{11} + w_{00} + w_{10} + w_{01}) \tau^2}
= w_{11} + w_{00} - w_{10} - w_{01}.
\]

Since $w_{10} + w_{01} > 0$ for $\rho_{34} < 1$ (the bivariate normal always 
assigns positive probability to discordant quadrants), 
$w_{11} + w_{00} - w_{10} - w_{01} < 1$.

Therefore, $\bar{\rho} = \sup_{\gamma} \rho^{\mathrm{fMCEM}}(\gamma) < 1$.
\end{proof}

\subsection{Proof of Proposition 3}
Proposition 3 is trivial from the law of total covariance in Remark 4. 
Hence we omit the proof here. 

\section{Additional simulation details}
\subsection{Data generation details}
We generate paired synthetic OU UWF image covariates $\mathcal{X}_1, \mathcal{X}_2 \in \mathbb{R}^{72 \times 72}$. 
Both images are partitioned into nine $24 \times 24$ square block matrices $\mathcal{X}^{(j)}_{t,s}$ 
such that
The synthetic images are generated as $\mathcal{X}_1, \mathcal{X}_2 \in \mathbb{R}^{72 \times 72}$ such that for $j=1, 2$, 
\[
\mathcal{X}_j =
\begin{pmatrix}
X_{j,1,1} & X_{j, 1,2} & X_{j, 1,3} \\
X_{j, 2,1} & X_{j, 2,2} & X_{j, 2,3} \\
X_{j, 3,1} & X_{j, 3,2} & X_{j, 3,3}
\end{pmatrix}, \quad
X_{j, t,s} \in \mathbb{R}^{24 \times 24} := X_{j, t,s}^{(k, l)}, ~ t, s = 1, 2, 3, ~k, l=1, \ldots, 24.
\]

The proximity between the OU UWF fundus images are characterized via pixel-wise joint normal distributions. 
Specifically, for pixels within the bottom-right blocks $(t,s) = (3,3)$, the pixel values are drawn as
$$ (\mathcal{X}_{1}^{(k,l)}, \mathcal{X}_{2}^{(k,l)})^T \sim N\left( \begin{pmatrix} 0.5 \\ 0.5 \end{pmatrix}, \begin{pmatrix} 0.25 & 0.125 \\ 0.125 & 0.25 \end{pmatrix} \right). $$
For pixels in all other blocks $(t,s) \neq (3,3)$, the values are generated as:
$$ (\mathcal{X}_{1}^{(k,l}, \mathcal{X}_{2}^{(k,l)})^T \sim MVN\left( \begin{pmatrix} 0 \\ 0 \end{pmatrix}, \begin{pmatrix} 0.5 & 0.25 \\ 0.25 & 0.5 \end{pmatrix} \right). $$
Based on these paired image covariates, we generate continuous responses (AL) $y_{1}, y_{3} \in \mathbb{R}$ and binary responses (myopia status) $y_{2}, y_{4} \in \{0, 1\}$. 
We define three specific spatial index sets to extract feature patches from the images: $\Omega_1 = \{1, 2, 3\}^2$, $\Omega_2 = \{4, 5, 6\}^2$, and $\Omega_3 = \{7, 8, 9\}^2$. For a given image $\mathcal{X}$, we construct the following block operators:
$$ S_1(\mathcal{X}) = \sum_{(t,s) \in \Omega_1} \tanh(\mathcal{X}^{(t,s)}), \quad S_2(\mathcal{X}) = \sum_{(t,s) \in \Omega_2} \mathcal{X}^{(t,s)}, \quad S_3(\mathcal{X}) = \sum_{(t,s) \in \Omega_3} \tanh(\mathcal{X}^{(t,s)}). $$
The nonlinear regression functions for  the two continuous responses $y_1$ and $y_3$ are formulated as $g(\mathcal{X}_j) = S_1(\mathcal{X}_j) + S_2(\mathcal{X}_j) + S_3(\mathcal{X}_j)$;  
for the binary responses $y_2$ and $y_4$, the regression functions are defined solely by the linear patch operator such that $g_{2}(\mathcal{X}_j) = S_2(\mathcal{X}_j)$, for $j=1, 2$. 


\subsection{Toy Example: Computational Efficiency of fMCEM}
\label{supp:toy}

We conduct a toy simulation study to compare the proposed fast
Monte Carlo Expectation Maximization (fMCEM) algorithm against
the extended rank-likelihood Markov Chain Monte Carlo (MCMC) approach
\citep{hoff2007extending} in terms of both estimation accuracy and
computational efficiency for estimating the copula correlation
matrix~$\Gamma$.

\paragraph{Data-generating mechanism.}
We generate $n = 800$ observations from a four-dimensional Gaussian
copula model with two continuous responses $(y_1, y_3)$ and two
binary responses $(y_2, y_4)$, following the notation of the main
text. The true correlation matrix is
\begin{equation}\label{eq:gamma_true_toy}
  \Gamma_{\mathrm{true}} =
  \begin{pmatrix}
    1     & 0.720 & 0.294 & 0.213 \\
    0.720 & 1     & 0.205 & 0.336 \\
    0.294 & 0.205 & 1     & 0.569 \\
    0.213 & 0.336 & 0.569 & 1
  \end{pmatrix}.
\end{equation}
The continuous responses are generated as
$y_k = \mathbf{X}_k^\top \boldsymbol{\beta} + \epsilon_k$
and the binary responses via a latent logistic threshold
$y_k = \mathbf{1}\{\mathbf{X}_k^\top \boldsymbol{\beta}' + u_k > 0\}$,
where $(\epsilon_1, \epsilon_2, u_1, u_2)^\top$ are drawn from the
Gaussian copula with $\Gamma_{\mathrm{true}}$ and the binary margins
use the probit--logistic link
$u_k = \Phi^{-1} \circ \mathcal{S}(v_k)$ with
$v_k \sim \mathcal{LO}$.
The covariates $\mathbf{X}_k \in \mathbb{R}^3$ are drawn from a
multivariate normal with AR(1) covariance $\sigma_{ij} =
0.75^{|i-j|}$.
After generating the data, plug-in inputs to both algorithms are
obtained from an ordinary least-squares fit (continuous responses)
and a logistic regression fit (binary responses), mimicking the
warm-up stage of the feasible CeViT framework (Algorithm~3 of the
main text).

\begin{figure}[!htb]
  \centering
  \includegraphics[width=\linewidth]{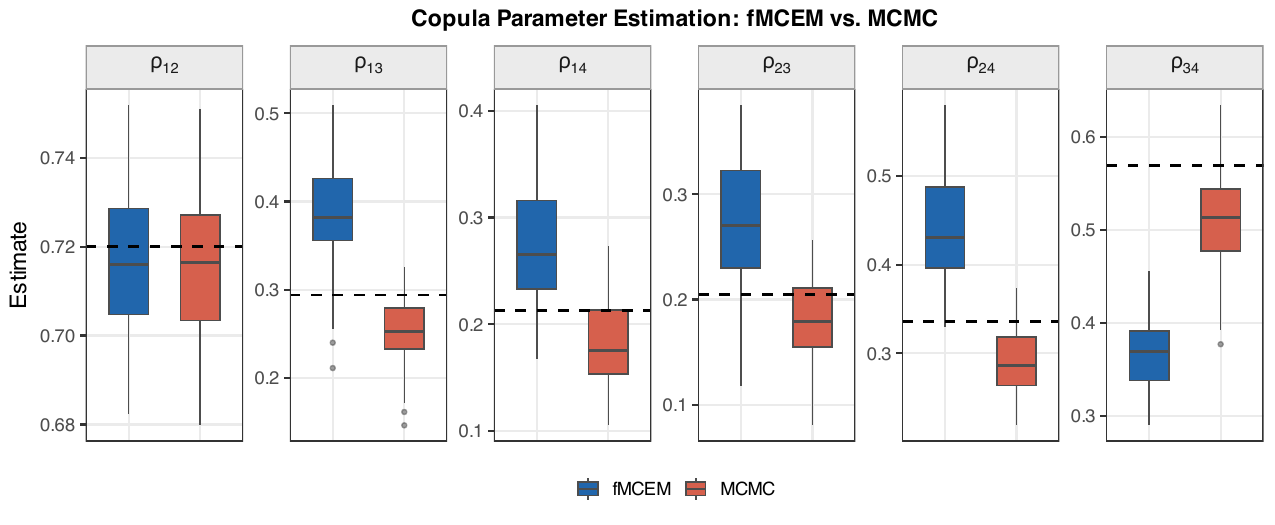}
  \caption{Boxplots of estimated copula correlation parameters over
           100 simulation replications ($n = 800$). Dashed horizontal
           lines indicate true values. Blue: fMCEM; Red: MCMC
           (coordinate-wise Gibbs sampler with inverse-Wishart update).}
  \label{fig:toy_params}
\end{figure}

\paragraph{Algorithm settings.}
The fMCEM algorithm (Algorithm~2) is run with a maximum of
$T = 100$ iterations and a convergence tolerance of $10^{-6}$ on
the Frobenius norm of successive correlation matrix updates.
The MCMC sampler is run for $2500$ total iterations with a burn-in of
$1000$ with a thinning of $10$. 
 The simulation is repeated for $B = 100$ independent replications 
\paragraph{Results: copula parameter estimation.}
Figure~\ref{fig:toy_params} displays boxplots of the estimated
off-diagonal elements of $\Gamma$ over 100 replications, with true
values indicated by dashed horizontal lines. 

The fMCEM algorithm recovers $\rho_{12}$ (the
continuous--continuous correlation) with negligible bias
($\mathrm{bias} = {-0.003}$, $\mathrm{SD} = 0.016$), matching
the MCMC estimator closely.
For the mixed continuous--binary correlations
$\rho_{13}, \rho_{14}, \rho_{23}, \rho_{24}$, fMCEM exhibits a
moderate positive bias (ranging from $+0.060$ to $+0.104$), while
MCMC shows a small negative bias (from $-0.024$ to $-0.046$).
For the binary--binary correlation $\rho_{34}$, fMCEM underestimates
the true value ($\mathrm{bias} = {-0.201}$, $\mathrm{SD} = 0.038$),
whereas MCMC is closer to the truth
($\mathrm{bias} = {-0.057}$, $\mathrm{SD} = 0.051$).
Such underestimation of $\rho_{34}$ is an expected and acknowledged
consequence of the approximation $\Sigma_{22i} \approx \tilde{V}$ in
the fMCEM derivation (Remark~3 of the main text). 

\begin{figure}[!htb]
  \centering
  \includegraphics[width=0.38\linewidth]{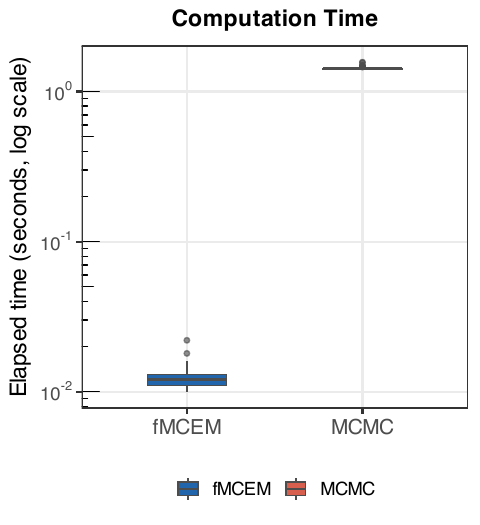}
  \caption{Boxplots of elapsed computation time (seconds,
           $\log_{10}$ scale) over 100 simulation replications.
           fMCEM is approximately $116$ times faster than the MCMC
           sampler.}
  \label{fig:toy_time}
\end{figure}

\paragraph{Results: computational efficiency.}
Figure~\ref{fig:toy_time} displays boxplots of elapsed computation
time (seconds, log scale) over 100 replications.
Table~\ref{tab:toy_time} summarises the key statistics.
We find that the fMCEM algorithm converges in all $100$ replications
(convergence rate $= 100\%$) within a mean of $24.0$ iterations
($\mathrm{SD} = 1.4$), with a mean elapsed time of $0.012$ seconds
($\mathrm{SD} = 0.002$ seconds).
The resulting speed-up factor is approximately $\mathbf{116}$-fold compared to the MCMC approach.
We note that as data size increases, the computational efficiency of the MCMC approach will linearly decrease. 
Therefore, the proposed fMCEM algorithm will perform higher efficiency on larger data. 

\begin{table}[ht]
  \centering
  \caption{Computational efficiency comparison over 100 simulation
           replications ($n = 800$). The MCMC sampler uses
           coordinate-wise Gibbs sampling via \texttt{truncnorm}.}
  \label{tab:toy_time}
  \begin{tabular}{lcccc}
    \hline
    Method & Mean time (s) & SD time (s) & Convergence & Mean iterations \\
    \hline
    fMCEM  & $0.012$       & $0.002$     & $100/100$   & $24.0\ (\pm 1.4)$ \\
    MCMC   & $1.420$       & $0.029$     & ---         & $2500$ (fixed) \\
    \hline
    \multicolumn{2}{l}{Time cost ratio (MCMC / fMCEM)} &
    \multicolumn{3}{c}{${\approx}116\times$} \\
    \hline
  \end{tabular}
\end{table}

\bibliographystyle{apalike}
\selectfont
\bibliography{ref}